\documentclass{article}
\usepackage{graphicx} 
\usepackage{amsmath}
\usepackage{amssymb}
\usepackage{dsfont}
\usepackage{subcaption}
\usepackage{xcolor}
\usepackage{graphicx}
\usepackage{wrapfig}
\definecolor{refpurple}{HTML}{6C569F}

\usepackage[
  colorlinks=true,
  linkcolor=refpurple, 
  citecolor=refpurple, 
  urlcolor=refpurple   
]{hyperref}
\definecolor{tbpblue}{HTML}{6C569F}
\newcommand{\tbp}[1]{\textcolor{tbpblue}{#1}}

\usepackage{booktabs}
\usepackage{makecell}
\usepackage{adjustbox}
\usepackage[table]{xcolor}
\usepackage{caption}
\usepackage{float}

\usepackage{titletoc}

\hypersetup{
  colorlinks=true,
  linkcolor=annpurple,
  citecolor=annpurple,
  urlcolor=annpurple,
  filecolor=annpurple
}
\definecolor{tablerule}{HTML}{D6DCE3}
\definecolor{pmgray}{HTML}{7D8B99}
\definecolor{scodabg}{HTML}{FCFAFE}
\definecolor{tbppurple}{HTML}{6C569F}
\definecolor{metricgreen}{HTML}{2E8B57}
\definecolor{metricred}{HTML}{C0392B}
\providecommand{\tbp}[1]{{\color{tbppurple}\bfseries #1}}

\usepackage[table]{xcolor}
\usepackage{booktabs}

\definecolor{scodaViolet}{HTML}{A885F2}
\definecolor{scodaVioletTint}{HTML}{F1E8FF}

\newcommand{\thcv}[1]{\cellcolor{scodaVioletTint}\textbf{#1}}

\newcommand{\vpm}[2]{\ensuremath{#1\,{\color{pmgray}\scriptstyle\pm\,#2}}}
\newcommand{\NA}{\textcolor{pmgray}{--}}
\newcommand{\metricup}{\textcolor{metricgreen}{\ensuremath{\uparrow}}}
\newcommand{\metricdown}{\textcolor{metricred}{\ensuremath{\downarrow}}}
\definecolor{Blue600}{HTML}{2563EB}

\usepackage[preprint]{corl_2026} 

\title{Task-Aware Environment Augmentation for Reliable Navigation via Shielded Conditional Diffusion}

\author{
Bharawee Phoompho, \quad
Gokul Puthumanaillam, \quad
Yan Miao, \\
\textbf{Ruben Hernandez}, \quad
\textbf{Tim Bretl}, \quad
\textbf{Sayan Mitra}, \quad
\textbf{Melkior Ornik} \\
\vspace{0.1cm}
\footnotesize  University of Illinois Urbana\hbox{-}Champaign (\texttt{\{bp16, gokulp2, mornik\}@illinois.edu})} 

\begin{document}
\maketitle

\begin{abstract}
 Reliable trajectory planning under partial observability depends not only on computing a feasible geometric path, but also on whether the robot receives informative observations while executing that trajectory. Existing approaches usually keep the environment fixed and adapt the robot through belief-space planning, active localization, or added sensing, often incurring costly uncertainty propagation and brittle behavior in observation-poor regions. We flip this perspective and address the largely open problem of \emph{task-aware environment augmentation}: given a mapped environment, a planned task trajectory, and a small budget of visual fiducial markers, where should the environment be augmented so that the planned trajectory can be executed reliably under uncertainty? Our key observation is that useful marker layouts are defined by the localization support they provide along the task trajectory: a small number of well-timed observations can be sufficient to prevent uncertainty from accumulating in regions where state-estimation error would otherwise compromise control. Building on this observation, we present \tbp{SCoDA}, $\textbf{S}$hielded $\textbf{Co}$nditional $\textbf{D}$iffusion for Environment $\textbf{A}$ugmentation. \tbp{SCoDA} learns a conditional distribution over high-performing fiducial layouts from data, using the environment, planned trajectory, disturbance context, and desired execution profile as conditioning. Its shielded sampler reasons over where along the planned execution pose corrections should occur, and steers this distribution toward task-relevant, finite-budget augmentations. Across simulated benchmarks and hardware deployments, we show that \tbp{SCoDA} improves trajectory execution reliability and completion time over strong baselines.
 Code, models and dataset available at: \hyperlink{scoda-diffusion.github.io}{https://scoda-diffusion.github.io/}
\end{abstract}
\keywords{Environment augmentation, Trajectory Planning, Diffusion models}

\section{Introduction}
\label{sec:introduction}

Autonomous robots executing navigation tasks under partial observability must solve two coupled problems: they must know where to go, and they must remain localized well enough to get there. In GPS-denied or perceptually degraded environments, a planned trajectory can become unreliable if the robot receives too few informative observations while executing it. Between observations, pose error grows under process noise, sensing uncertainty, and external disturbances; once the estimate drifts far enough, even a good controller may track the state incorrectly and fail the task. The standard response is to make the robot more adaptive through richer sensing, belief-space planning, or active localization. We study the complementary question: given a mapped environment and a trajectory-planning task, where should a small number of visual fiducials be added so the planned trajectory receives \textit{just enough} localization support to be executed reliably?

Prior work largely treats limited observability as a robot-side problem: localization and SLAM estimate from whatever measurements are available~\cite{thrun2005probabilistic,fox1999montecarlo,durrantwhyte2006slam,cadena2016past}, while belief-space and active-localization methods adapt motion to manage or seek information~\cite{kaelbling2013integrated,indelman2015planning,platt2010belief,van2012lqgmp,burgard1997active,thrun1998active}. In many robotics deployments, the environment can be instrumented for reliable execution. 
For known trajectories in warehouses, inspection corridors, indoor flight arenas, or factory floors, placing a small number of visual landmarks can be less intrusive than adding sensing hardware, replanning online, or forcing localization-seeking detours.

Visual fiducials such as AprilTags and ArUco markers provide a practical way to add known visual reference points to a mapped environment~\cite{olson2011apriltag,wang2016apriltag2,garrido2014aruco,kalaitzakis2021fiducial}. When visible, they provide pose measurements that help the onboard estimator correct accumulated state error, so their placement determines where along the trajectory localization support is available. However, uniform or manual placements can waste fiducials in regions where the estimate is already reliable, and pure visibility objectives only measure whether a fiducial can be seen, not whether it improves execution. Furthermore, dense augmentation is costly, visually intrusive, and often unnecessary \cite{beinhofer2013effective, magnago2019effective}. Therefore, we seek a budgeted, task-specific augmentation: adding \textit{just enough} visual structure to keep the robot localized during the given task. Choosing where to place this sparse visual structure requires reasoning about the interaction between the planned trajectory, sensing model, disturbances, estimator, and controller, motivating a data-driven model learned from closed-loop rollout outcomes.


We formulate sparse environment augmentation as conditional generative design: learning a distribution over sparse fiducial placements that make a prescribed trajectory reliable under uncertainty. The key observation is that the value of a fiducial is not determined by visibility alone, but by its closed-loop effect on execution: a detection is useful when it corrects the state estimate early enough to improve tracking of the planned trajectory. This makes placement depend on the environment, reference trajectory, disturbances, estimator, and controller, rather than on a static coverage objective. Building on diffusion models for multimodal structured generation~\cite{sohl2015deep,ho2020denoising,song2021score}, we introduce \tbp{SCoDA}, Shielded Conditional Diffusion for Environment Augmentation.

\begin{figure}[h]
    \centering
    \includegraphics[width=1\linewidth]{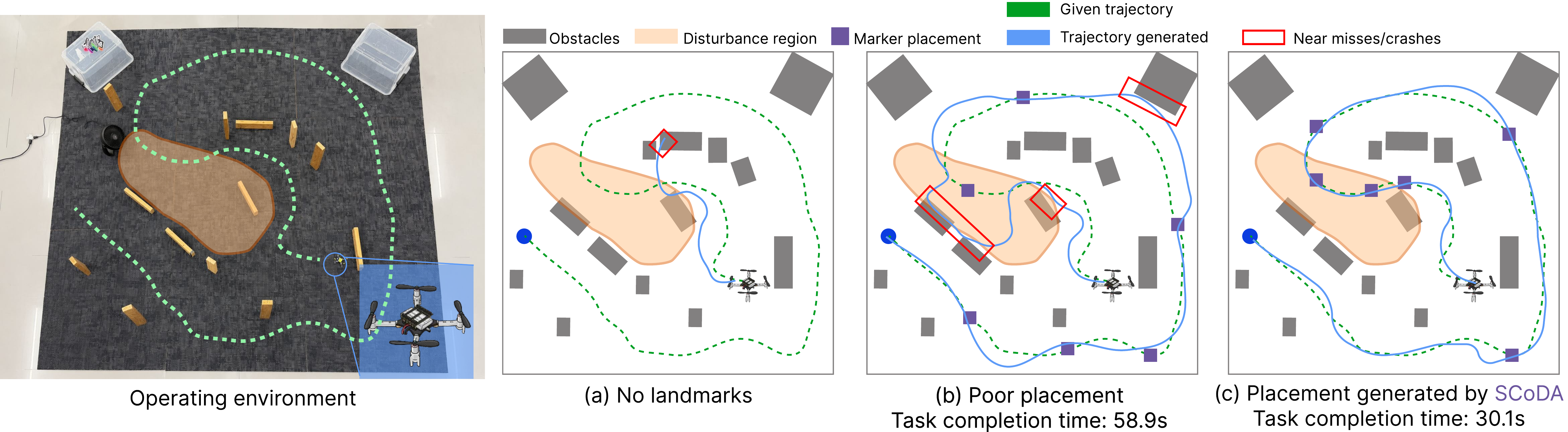}
    \caption{Task-aware fiducial augmentation. Poor or missing fiducial support can cause trajectory-tracking failures, while \tbp{SCoDA} places sparse fiducials where localization support improves execution.}
    \label{fig:placement_examples}
\end{figure}
\tbp{SCoDA} learns from closed-loop rollout data by pairing fiducial placements with their execution outcomes. Given a new task, it generates fiducial placements conditioned on the environment, reference trajectory, disturbance context, and desired execution profile. Its shielded reverse sampler uses the same task context during inference to steer the learned distribution toward finite-budget placements whose detections are likely to affect execution, while avoiding redundant placements. 
Across simulated benchmarks and real-time hardware experiments, \tbp{SCoDA} improves trajectory execution reliability and completion time over heuristic and unguided placement baselines.
To spur further research in this domain, we release the code, models, and dataset on the project website.


\section{Related Work}
\label{sec:related_work}

\textit{Probabilistic localization and belief-space planning.}
Robot navigation under uncertainty is commonly formulated in terms of belief
states, where the robot maintains a distribution over possible poses rather than
a single deterministic state~\cite{thrun2005probabilistic,fox1999montecarlo}.
Classical SLAM and localization algorithms estimate this belief through probabilistic
filtering or factorized map representations~\cite{dissanayake2001solution,montemerlo2002fastslam}.
Active localization and active perception methods choose actions to reduce pose
uncertainty or increase future observation informativeness
~\cite{burgard1997active,thrun1998active,bajcsy2018active, liberzon2025indistinguishability,yuceel2026active}. Belief-space
planners reason about future uncertainty through covariance propagation,
maximum-likelihood observations, sampling-based belief trees, or feedback
roadmaps~\cite{van2012lqgmp,platt2010belief,prentice2009belief,bry2011rrbt,aghamohammadi2014firm}.
Chance-constrained and Bayesian formulations further encode risk bounds in the
planner~\cite{censi2008bayesian,blackmore2011chance}. These methods primarily
optimize robot actions or trajectories; our work instead modifies the environment
by placing visual landmarks so a reference trajectory can be followed more
reliably.

\textit{Informative planning and landmark-based localization.}
Information-theoretic planning improves mapping, localization, or exploration by selecting trajectories that provide informative observations \cite{bourgault2002information,stachniss2005information,charrow2015information,hollinger2014sampling}. In camera-based navigation, these observations often depend on stable visual
features, as shown by visual SLAM systems~\cite{davison2003realtime,strasdat2010scale}. Artificial fiducials
such as AprilTags and ArUco markers provide engineered features with known
identity and geometry for pose estimation
~\cite{olson2011apriltag,wang2016apriltag2,garrido2014aruco,kalaitzakis2021fiducial}.
Landmark placement has been studied directly: landmark selection in natural
terrain~\cite{olson2002selecting}, geometric landmark distinguishability~\cite{erickson2012artgallery}, placement for reliable mobile robot
navigation~\cite{beinhofer2013effective}, indoor localization with uncertainty
constraints~\cite{magnago2019effective}, and fiducial placement for visual
localization~\cite{huang2023optimizing}. These approaches typically use explicit
optimization, greedy selection, or hand-designed objectives, whereas we learn a
conditional generative model over landmark configurations.

\textit{Diffusion models and guided generation for planning and robotics.}
Diffusion models are generative models that sample data by reversing a gradual noising process
~\cite{sohl2015deep,ho2020denoising,song2021score}, and guidance can steer generation toward desired properties or constraints~\cite{dhariwal2021diffusion,ho2022classifierfree,bansal2024universal}. Their ability to model multimodal distributions has made them useful for generating robot trajectories, actions, and manipulation behaviors~\cite{janner2022diffuser,ajay2023decisiondiffuser,carvalho2023motion,chi2023diffusionpolicy,urain2023se3diffusionfields,mishra2023reorientdiff}. Diffusion has also been combined with cost-guided sampling for autonomous driving and planning~\cite{yang2024diffusiones}. In contrast, \tbp{SCoDA} uses diffusion to generate environment augmentations: sparse fiducial layouts that provide localization support for a prescribed trajectory. Our guidance objective is specific to this setting, encouraging coverage of task-relevant trajectory intervals while discouraging redundant fiducial placements.

\section{Problem Formulation}
\label{subsec:problem}

Consider a robot assigned a trajectory-planning task in a mapped environment $\mathcal{E}$ under partial observability. One example is an indoor mobile robot executing a planned trajectory through a warehouse aisle, inspection corridor, or flight arena, where GPS is unavailable and camera-based localization corrections are intermittent. Let $\tau=(x^r_0,\ldots,x^r_N)$ denote the reference trajectory for the task. During execution, the robot does not observe its true state $x_t$ directly; it maintains an estimate $\hat{x}_t$ and applies a feedback controller $a_t=\pi(\hat{x}_t,\tau)$. The true state evolves as $x_{t+1}=f(x_t,a_t)+w_t$, where $w_t$ captures process noise and external disturbances. We write $\xi$ for the rollout randomness, including process noise, measurement noise, initial estimation error, and disturbance realizations.

Before execution, the planner may augment the environment with at most $K$ visual fiducials. A visual fiducial is a calibrated marker with known identity and pose; when detected by the robot's camera, it provides a measurement $z_t$ that the robot's existing estimator can use to correct $\hat{x}_t$. We denote a candidate augmentation by $\mathbf{m}\in\mathcal{M}_K$, where $\mathbf{m}$ is a feasible layout of at most $K$ physical fiducial poses in the mapped environment. 

Let $\mathcal{V}(x_t)$ denote the set of fiducial poses detectable from robot state $x_t$ under the camera model, including range, field of view, occlusion, and marker-orientation constraints. A fiducial correction is available when $\alpha_t(\mathbf{m})=\mathds{1}[\mathbf{m}\cap\mathcal{V}(x_t)\neq\emptyset]$. If $\alpha_t(\mathbf{m})=1$, the estimator incorporates the detected fiducial measurement $z_t$; if $\alpha_t(\mathbf{m})=0$, the estimate is propagated using the motion model without a fiducial correction. Thus, the chosen augmentation changes the closed-loop execution by changing when fiducial-based pose corrections are available along the planned trajectory.

The objective is to choose a sparse augmentation that makes the entire trajectory execution reliable. Let $e_t^\tau$ denote the tracking error between the executed state and the reference trajectory at the corresponding progress along $\tau$, and let $t_f$ be the first time the terminal task region is reached. Let $\mathcal{F}_\tau$ denote the event that trajectory execution fails, for example by violating task tolerances or not reaching the terminal region within the horizon $T_{\max}$. The rollout randomness follows $\xi\sim p(\xi)$, the matrix $Q\succeq0$ weights tracking error, the scalar $\lambda_{\mathrm{time}}$ weights completion time, and $C_{\mathrm{fail}}$ is a large penalty assigned to failed executions. We define the budgeted augmentation objective as
\begin{equation}
\label{eq:placement_problem}
\mathbf{m}^{\star}\in\arg\min_{\mathbf{m}\in\mathcal{M}_K}\mathbb{E}_{\xi\sim p(\xi)}\left[\sum_{t=0}^{T_{\max}}\|e_t^\tau\|_Q^2+\lambda_{\mathrm{time}} t_f+C_{\mathrm{fail}}\mathbb{I}[\mathcal{F}_\tau]\right].
\end{equation}

This objective favors augmentations that keep the robot localized well enough to track the planned trajectory, reach the terminal task region efficiently, and avoid execution failure under stochastic disturbances. 
Solving Eq.~\eqref{eq:placement_problem} directly for every new task would require repeated closed-loop rollouts over candidate augmentations and would typically return only one layout. We instead learn a conditional generator that produces high-performing, task-specific augmentation candidates from the environment, reference trajectory, disturbance context, and execution statistics.

\section{\tbp{SCoDA}: Shielded Conditional Diffusion for Environment Augmentation}
\label{subsec:conditional_diffusion}

\label{subsec:scoda}

\tbp{SCoDA} is an amortized solver for the budgeted augmentation problem in Eq.~\eqref{eq:placement_problem}. Instead of running a new search over fiducial placements for every trajectory-planning task, \tbp{SCoDA} learns to generate sparse augmentations that have supported reliable closed-loop execution in similar task contexts. The method has three main components: path-coordinate fiducial representation, rollout-conditioned diffusion training, and shielded reverse diffusion to cover task-critical regions while avoiding redundant tag placements.

\begin{figure}[h]
    \centering
    \includegraphics[width=1\linewidth]{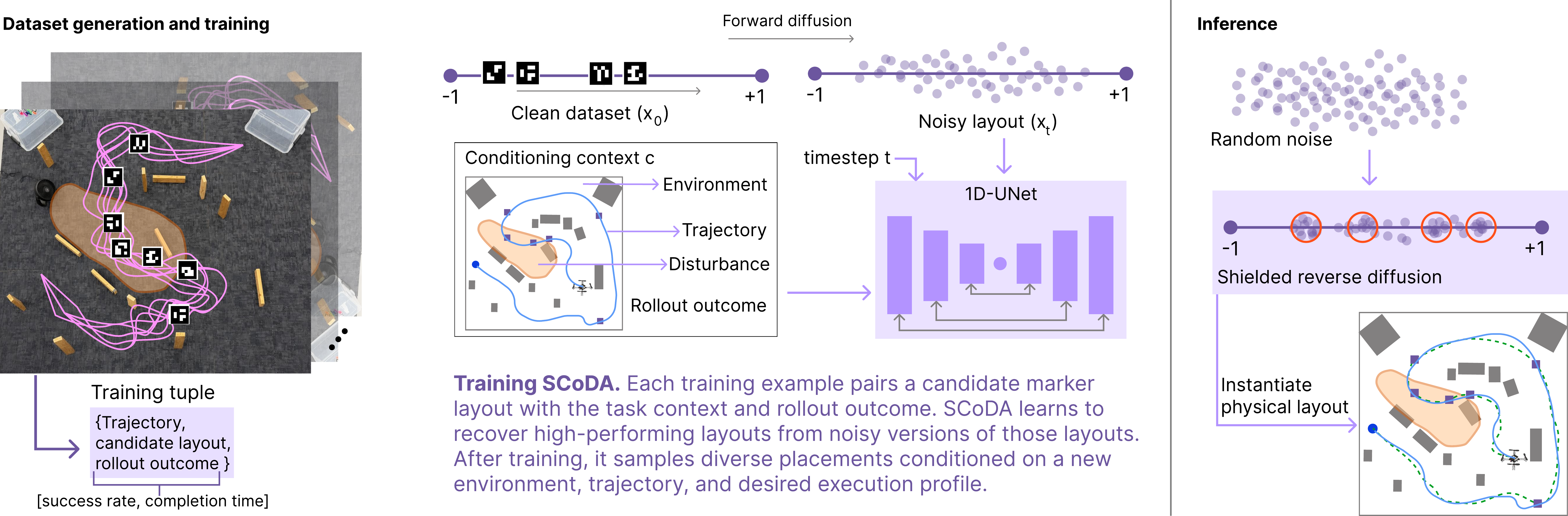}
    \caption{\tbp{SCoDA} pipeline: a conditional diffusion model iteratively denoiseDs landmark positions in trajectory-progress space, guided by task context and shielded inference.}
    \label{fig:diagram}
\end{figure}

\paragraph{Trajectory-indexed augmentation.}
\begin{wrapfigure}[14]{r}{0.60\textwidth}
    \vspace{-1.5em}
    \centering
\includegraphics[width=0.58\textwidth]{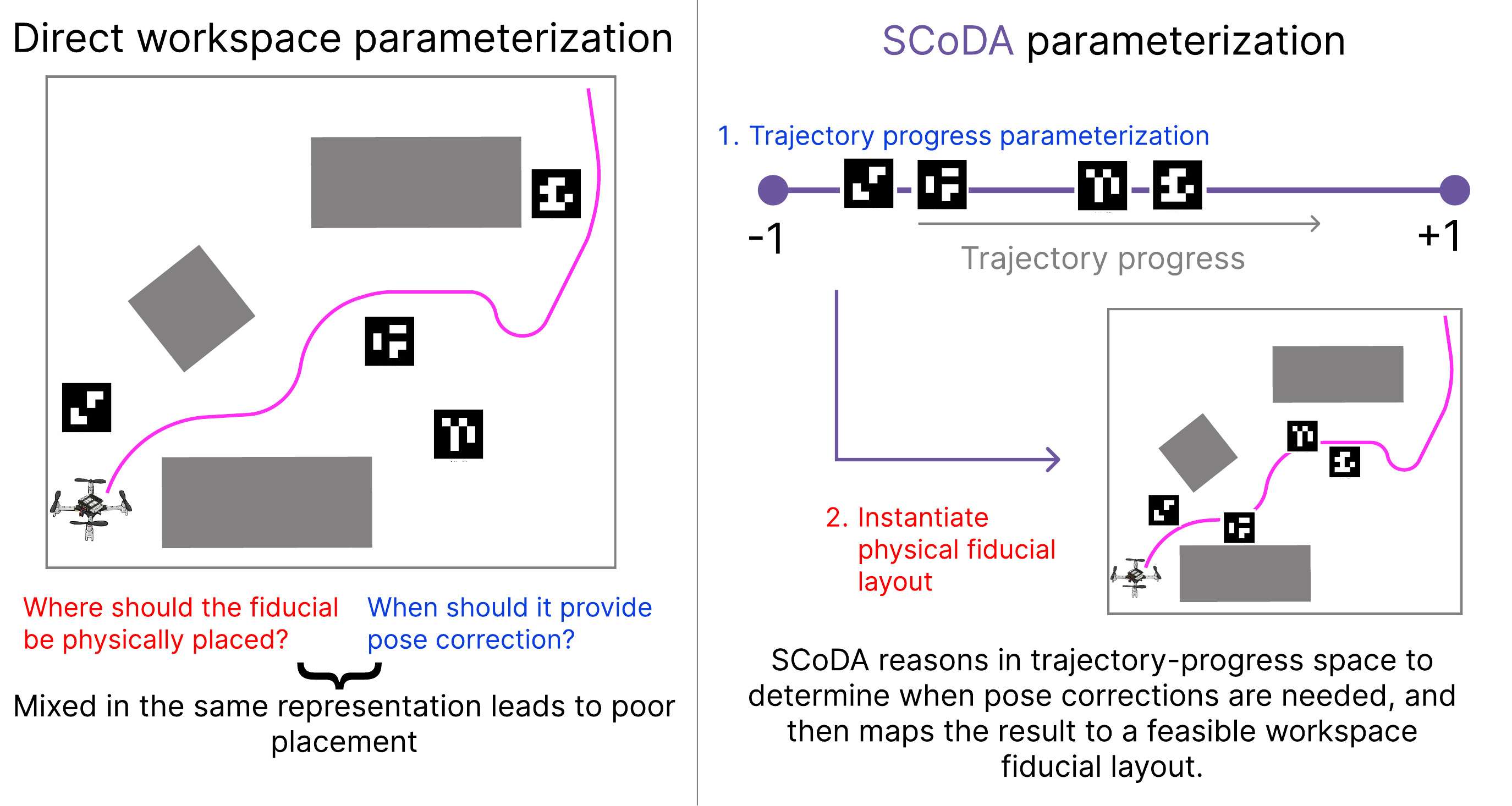}
    \caption{\tbp{SCoDA} decouples when a pose correction is needed from where the fiducial is placed by reasoning in trajectory-progress space before mapping back to a physical layout.}
    \label{fig:pl1}
\end{wrapfigure}
A generative model for environment augmentation must decide how to describe the candidate layout $\mathbf{m}$. Generating fiducial poses directly in workspace coordinates forces the model to learn two things at once: where a fiducial should physically be placed, and where along the planned execution its detection should correct the state estimate. \tbp{SCoDA} therefore generates an intermediate progress-space variable $\mathbf{u}\in[-1,1]^K$, where each $u_i$ specifies the point along $\tau$ at which the corresponding fiducial should provide a pose correction (see Figure~\ref{fig:pl1}). After sampling, $\mathbf{u}$ is instantiated as a feasible physical layout $\mathbf{m}\in\mathcal{M}_K$ in the mapped environment. This parameterization keeps generation tied to execution: task-relevant parts of $\tau$ become intervals over $\mathbf{u}$, and redundant fiducials correspond to progress values that are too close together.

\paragraph{Rollout-conditioned supervision.}
\tbp{SCoDA} learns from the effect an augmentation has on closed-loop trajectory execution. For each training task, an augmentation vector $\mathbf{u}$ is mapped to fiducial poses through $M_{\mathcal{E},\tau}$, and the robot executes the reference trajectory over repeated stochastic rollouts with the task controller, estimator, process noise, measurement noise, and disturbance model. The resulting execution statistics define $\mathbf{y}=[s_r,\bar{t}_c]$, where $s_r$ is the fraction of successful trajectory executions and $\bar{t}_c$ is the mean completion time normalized by $T_{\max}$. The conditional input is $\mathbf{c}=\left[g_x,g_y,\psi(\tau),\mathbf{d},\mathbf{y}\right],$
where $(g_x,g_y)$ is the terminal task location, $\psi(\tau)$ is a fixed-size trajectory encoding obtained by uniformly sampling the reference trajectory in progress, and $\mathbf{d}=[u_{d,s},u_{d,e},u_{r,e}]$ specifies task-relevant intervals along the trajectory. Here, $u_{d,s}$ and $u_{d,e}$ mark the start and end of the disturbance interval, while $u_{r,e}$ is the end of the subsequent recovery interval, where additional localization support may be useful. During training, $\mathbf{y}$ is computed from the rollout outcomes of each candidate layout. At inference, we set $\mathbf{y}$ to the target profile $\mathbf{y}^{\star}=[1,\bar{t}_{\mathrm{des}}]$, querying the model for augmentations associated with reliable and efficient execution.

\paragraph{Diffusion model over sparse augmentations.}
For a fixed context $\mathbf{c}$, there may be more than one augmentation vector $\mathbf{u}$ that leads to reliable execution. 
We therefore learn a conditional distribution $p_\theta(\mathbf{x}_0\mid\mathbf{c})$ over augmentation vectors, where $\mathbf{x}_0$ is the diffusion input corresponding to $\mathbf{u}$. Following \cite{ho2020denoising}, a clean augmentation vector is corrupted as $q(\mathbf{x}_t\mid\mathbf{x}_0)=\mathcal{N}(\sqrt{\bar{\alpha}_t}\mathbf{x}_0,(1-\bar{\alpha}_t)\mathbf{I})$, and the denoiser is trained to predict the injected noise,
\begin{equation}
    \mathcal{L}_{\mathrm{diff}}=\mathbb{E}_{\mathbf{x}_0,\mathbf{c},t,\epsilon}\left[\left\|\epsilon-\epsilon_\theta(\mathbf{x}_t,t,\tilde{\mathbf{c}})\right\|_2^2\right],
    \label{eq:scoda_diffusion_loss}
\end{equation}
where $\epsilon\sim\mathcal{N}(\mathbf{0},\mathbf{I})$ and $\tilde{\mathbf{c}}$ is set to a null context with fixed probability during training. This conditioning dropout lets the same network produce both conditional and unconditional predictions. At inference, we combine the two predictions using classifier-free guidance \cite{Ho2021ClassifierFreeGuidance},
\begin{equation}
    \hat{\epsilon}_{\mathrm{cfg}}=\epsilon_\theta(\mathbf{x}_t,t,\mathbf{0})+w_{\mathrm{cfg}}\left(\epsilon_\theta(\mathbf{x}_t,t,\mathbf{c})-\epsilon_\theta(\mathbf{x}_t,t,\mathbf{0})\right).
    \label{eq:scoda_cfg}
\end{equation}
The guidance weight $w_{\mathrm{cfg}}$ controls how strongly generation follows the requested context and target execution profile. Starting the reverse process from different noise samples gives multiple candidate fiducial layouts for the same trajectory-planning task. The layouts are then ranked by the placement objective or closed-loop evaluation before returning a single selected layout.

\paragraph{Shielded reverse diffusion.}
Rollout-conditioned diffusion biases samples toward layouts that performed well in the training data, but it does not ensure that every generated layout uses the finite tag budget effectively. In particular, a sample may cluster several tags at nearly the same trajectory progress or miss an interval where the context indicates that a correction is needed. We therefore add a generation-time shield: a differentiable guidance objective that steers samples toward simple structural requirements without rerunning closed-loop rollouts during denoising. This differentiable modification of the denoising update follows guided diffusion methods, which steer sampling with classifier gradients, classifier-free scores, or external objectives \cite{dhariwal2021diffusion,ho2022classifierfree}. Formally, this corresponds to sampling from a guided distribution
\begin{equation}
q_\lambda(\mathbf{u}\mid\mathbf{c})\propto p_\theta(\mathbf{u}\mid\mathbf{c})\exp\left(-\lambda J_{\mathrm{task}}(\mathbf{u};\mathbf{c})\right),
    \label{eq:scoda_tilt}
\end{equation}
where $J_{\mathrm{task}}$ assigns low cost to layouts that satisfy the task-level placement preferences and $\lambda$ controls the strength of shielding. This distributional view gives a useful sanity check: under exact sampling from Eq.~\eqref{eq:scoda_tilt}, $\frac{d}{d\lambda}\mathbb{E}_{q_\lambda}[J_{\mathrm{task}}]=-\mathrm{Var}_{q_\lambda}(J_{\mathrm{task}})\leq 0$, so increasing $\lambda$ monotonically lowers the expected task-objective violation. The derivation is given in {Appendix~\ref{proofs}}.

To implement this guidance during denoising, the current diffusion state is mapped to trajectory-progress coordinates $\mathbf{v}_t\in[0,1]^K$. Let $\mathcal{I}(\mathbf{c})=\{[a_m,b_m]\}_{m=1}^M$ denote the task-relevant trajectory intervals specified by the context. For an interval $[a,b]$, we use the smooth membership score $\psi(v;a,b)=\sigma(\kappa(v-a))\sigma(\kappa(b-v))$, which is close to 1 when $v$ lies inside the interval and close to 0 outside it. The task objective is
\begin{equation}
    J_{\mathrm{task}}(\mathbf{v}_t;\mathbf{c})=\sum_{m=1}^{M}\beta_m\left[1-\sum_{k=1}^{K}\psi(v_{t,k};a_m,b_m)\right]_+ + \beta_s\sum_{i<j}\sigma\left(\kappa_s(\delta_{\min}-|v_{t,i}-v_{t,j}|)\right).
    \label{eq:scoda_shield}
\end{equation}
The first term penalizes intervals that receive no fiducial-induced pose correction. The second term penalizes pairs of fiducials that are too close along trajectory progress, since nearby fiducials usually provide redundant corrections under a fixed budget. The task gradient is injected into the reverse update through the predicted noise, 
\begin{equation}
    \hat{\epsilon}_{\mathrm{sh}}=\hat{\epsilon}_{\mathrm{cfg}}+\lambda_t\nabla_{\mathbf{x}_t}J_{\mathrm{task}}(\mathbf{v}_t(\mathbf{x}_t);\mathbf{c}).
    \label{eq:scoda_shield_update}
\end{equation}
With the DDPM reverse-mean parameterization, this guidance moves the next sample toward lower $J_{\mathrm{task}}$ while the denoiser keeps it close to the rollout-conditioned distribution. Shielding is applied only over intermediate reverse steps, after noise-dominated early states and before late states become nearly fixed.

\paragraph{Candidate validation.}
After denoising, the generated vector is clipped to $[-1,1]^K$ and interpreted as the augmentation $\mathbf{u}$. The deployment map $M_{\mathcal{E},\tau}$ then maps $\mathbf{u}$ to physical fiducial poses in the environment. We optionally apply a final feasibility check using the same trajectory-interval and separation requirements encoded in $J_{\mathrm{task}}$. This check gives a direct guarantee on the returned layout: every retained augmentation satisfies the specified interval-coverage and minimum-separation constraints. If one generated sample passes the check with probability $p_{\mathrm{acc}}$, then $R$ independent samples yield at least one feasible augmentation with probability $1-(1-p_{\mathrm{acc}})^R$; see {Appendix~\ref{proofs}}. When closed-loop evaluation is available, multiple feasible candidates can be ranked by the objective in Eq.~\eqref{eq:placement_problem}; otherwise, the candidate with the lowest $J_{\mathrm{task}}$ is selected.

\paragraph{Implementation recipe.}
The denoiser is a one-dimensional U-Net with input and output dimension $K$.
Sinusoidal timestep embeddings are combined with a learned projection of the context vector $\mathbf{c}$ and injected into the residual blocks. Trajectory-progress values are represented in $[-1,1]$ for diffusion and rescaled to $[0,1]$ when evaluating $J_{\mathrm{task}}$; workspace quantities are normalized by the environment dimensions, and completion time is normalized by $T_{\max}$. The task objective uses smooth interval-membership and separation penalties, so the guidance gradient in Eq.~\eqref{eq:scoda_shield_update} is computed by automatic differentiation without modifying the denoising network. Diffusion-schedule parameters, classifier-free guidance weights, task-guidance weights, and feasibility-check thresholds are reported in {Appendix~\ref{proofs}}.
\section{Experimental Results}
\label{sec:experiments}

We evaluate \tbp{SCoDA} in simulation and hardware. In simulation, we use FalconGym 2.0 \cite{miao2026performanceguided}, a photorealistic aerial-navigation benchmark. We trivially modify it to support external disturbance wind fields. FalconGym's edit API lets us instantiate visual tags at generated locations while keeping the robot dynamics, rendering, and closed-loop execution fixed across methods~\cite{miao2025falcongym}. 
We evaluate two simulated environment settings: nominal, with no external perturbation, and disturbance-augmented, with localized wind fields along the reference trajectory.
The simulated vehicle uses FalconGym's fixed-wing aerial dynamics \cite{miao2025falconwing} and its default low-level tracking controller. For hardware, we deploy generated augmentations on a Crazyflie 2.0 quadrotor in a cluttered indoor environment with physical fiducial tags placed according to the generated layouts. All hardware trials use the onboard low-level flight controller and the same reference trajectory, estimator interface, and fiducial-detection pipeline across methods. Additional details on the setup are provided in {Appendix~\ref{expt}}.

We compare \tbp{SCoDA} against baselines chosen to isolate trajectory-indexed generation, rollout conditioning, shielding, and test-time optimization, while keeping the fiducial budget $K$, estimator, controller, and reference trajectory fixed. \textit{(i) No-Augmentation \textbf{(NA)}} uses no added fiducials. \textit{(ii) Random \textbf{(Rand)}} places $K$ fiducials at randomly sampled feasible locations. \textit{(iii) Periodic-$K$ \textbf{(PK)}} places $K$ fiducials at equidistant locations along the reference trajectory. \textit{(iv) Periodic-Dense \textbf{(PD)}} keeps adding periodic fiducials until the maximum observation gap falls below a threshold, giving an unbounded coverage baseline for measuring frugality. \textit{(v) Critical-Region \textbf{(CR)}} places fiducials near disturbance, recovery, and terminal-approach regions. \textit{(vi) Visibility-Greedy \textbf{(VG)}} greedily selects fiducials that make the largest number of reference states observable. \textit{(vii) Localizability-Greedy \textbf{(LG)}} selects fiducials using a camera-localizability score inspired by prior marker-placement objectives~\cite{huang2023optimizing}. \textit{(viii) Deviation-Greedy \textbf{(DG)}} places fiducials near regions with large predicted tracking or estimation-error growth~\cite{beinhofer2013effective} rather than optimizing for $[s_r, \bar{t}_c]$. \textit{(ix) Rollout-Opt \textbf{(RO)}} directly optimizes Eq.~\eqref{eq:placement_problem} with closed-loop rollouts and serves as an expensive upper bound. We also ablate \tbp{SCoDA} by removing shielding, rollout conditioning, disturbance context, trajectory-indexed generation, and shielded denoising. Baseline implementation details are provided in {Appendix~\ref{expt}}. 

\subsection{Results and Key Findings}

\paragraph{Key Finding \#1: \tbp{SCoDA} gives strong execution performance with just-enough augmentation.}
\begin{wraptable}{r}{0.4\textwidth}
\vspace{-1.25em}
\centering
\caption{
Hardware deployment results over 45 trials; $K=8$.
}
\label{tab:hardware}
{%
\scriptsize
\setlength{\tabcolsep}{2.6pt}
\renewcommand{\arraystretch}{1.14}
\arrayrulecolor{tablerule}
\begin{adjustbox}{max width=\linewidth,center}
\begin{tabular}{@{}lccc@{}}
\toprule
\textbf{Method}
& \makecell{\textbf{Success}\\\textbf{rate} (\%) \metricup}
& \makecell{\textbf{Waypoint}\\\textbf{following} (\%) \metricup}
& \makecell{\textbf{Completion}\\\textbf{time} (s) \metricdown} \\
\midrule

NA
& \vpm{4.6}{3.1}
& \vpm{24.2}{5.4}
& \vpm{58.4}{1.4} \\

Rand
& \vpm{13.7}{5.1}
& \vpm{36.9}{6.2}
& \vpm{55.6}{2.3} \\

PK
& \vpm{57.4}{7.4}
& \vpm{72.8}{6.1}
& \vpm{44.8}{3.4} \\

VG
& \vpm{68.7}{6.9}
& \vpm{78.9}{5.7}
& \vpm{40.9}{3.1} \\


\rowcolor{scodabg}
\tbp{SCoDA}
& \tbp{\vpm{91.2}{4.4}}
& \tbp{\vpm{94.7}{3.2}}
& \tbp{\vpm{29.5}{2.1}} \\

\bottomrule
\end{tabular}
\end{adjustbox}
\arrayrulecolor{black}
}%
\vspace{-0.75em}
\end{wraptable}
Across simulation and hardware, \tbp{SCoDA} is the strongest placement method under a fixed fiducial budget. Random and periodic layouts show that adding fiducials alone is insufficient; \tbp{SCoDA} improves over greedy methods by learning which observations actually support closed-loop execution. Table~\ref{tab:budget_efficiency} further shows that \tbp{SCoDA} reaches target thresholds with the smallest deployable budget, matching the rollout-optimized upper bound while saving markers over dense periodic coverage.


\begin{table}[H]
\centering

\begin{minipage}[t]{0.72\textwidth}
\centering
\caption{
Simulated trajectory execution with and without disturbances for $K=6$. Results are reported as mean $\pm$ standard error over 150 simulated rollouts.
}
\label{tab:sim_combined}
{%
\scriptsize
\setlength{\tabcolsep}{3.0pt}
\renewcommand{\arraystretch}{1.14}
\arrayrulecolor{tablerule}
\begin{adjustbox}{max width=\linewidth,center}
\begin{tabular}{@{}lcccccccc@{}}
\toprule
\textbf{Method}
& \multicolumn{4}{c}{\textbf{No external disturbances}}
& \multicolumn{4}{c}{\textbf{With disturbances}} \\
\cmidrule(lr){2-5}
\cmidrule(lr){6-9}
&
\makecell{\textbf{Success}\\\textbf{rate} (\%) \metricup}
&
\makecell{\textbf{Waypoint}\\\textbf{following} (\%) \metricup}
&
\makecell{\textbf{Tracking}\\\textbf{error} (m) \metricdown}
&
\makecell{\textbf{Completion}\\\textbf{time} (s) \metricdown}
&
\makecell{\textbf{Success}\\\textbf{rate} (\%) \metricup}
&
\makecell{\textbf{Waypoint}\\\textbf{following} (\%) \metricup}
&
\makecell{\textbf{Tracking}\\\textbf{error} (m) \metricdown}
&
\makecell{\textbf{Completion}\\\textbf{time} (s) \metricdown} \\
\midrule

NA
& \vpm{5.4}{1.8}
& \vpm{22.7}{3.9}
& \vpm{0.96}{0.07}
& \vpm{56.9}{1.2}
& \vpm{1.8}{1.1}
& \vpm{13.6}{2.8}
& \vpm{1.31}{0.10}
& \vpm{59.2}{0.8} \\

Rand
& \vpm{9.6}{2.4}
& \vpm{29.4}{4.6}
& \vpm{0.89}{0.08}
& \vpm{54.5}{1.5}
& \vpm{5.9}{1.9}
& \vpm{23.8}{3.9}
& \vpm{1.03}{0.08}
& \vpm{56.8}{1.2} \\

PK
& \vpm{64.8}{4.2}
& \vpm{76.3}{3.5}
& \vpm{0.34}{0.03}
& \vpm{39.2}{1.7}
& \vpm{43.7}{4.1}
& \vpm{58.9}{4.2}
& \vpm{0.56}{0.05}
& \vpm{47.4}{2.0} \\

CR
& \NA
& \NA
& \NA
& \NA
& \vpm{70.8}{3.7}
& \vpm{79.4}{3.2}
& \vpm{0.34}{0.03}
& \vpm{39.3}{1.7} \\

VG
& \vpm{72.6}{3.8}
& \vpm{82.1}{3.1}
& \vpm{0.29}{0.03}
& \vpm{35.4}{1.5}
& \vpm{57.9}{4.0}
& \vpm{68.1}{3.8}
& \vpm{0.45}{0.04}
& \vpm{44.0}{1.9} \\

LG
& \vpm{78.4}{3.4}
& \vpm{86.6}{2.7}
& \vpm{0.25}{0.02}
& \vpm{33.4}{1.4}
& \vpm{64.6}{3.9}
& \vpm{73.8}{3.5}
& \vpm{0.39}{0.04}
& \vpm{41.2}{1.8} \\

DG
& \vpm{84.1}{3.0}
& \vpm{89.7}{2.4}
& \vpm{0.21}{0.02}
& \vpm{31.2}{1.3}
& \vpm{77.3}{3.4}
& \vpm{84.9}{2.8}
& \vpm{0.28}{0.03}
& \vpm{35.2}{1.5} \\

RO
& \vpm{98.7}{0.9}
& \vpm{99.1}{0.6}
& \vpm{0.12}{0.01}
& \vpm{26.3}{0.7}
& \vpm{97.8}{1.1}
& \vpm{98.6}{0.7}
& \vpm{0.13}{0.01}
& \vpm{27.9}{0.8} \\

\rowcolor{scodabg}
\tbp{SCoDA}
& \tbp{\vpm{96.9}{1.3}}
& \tbp{\vpm{98.3}{0.8}}
& \tbp{\vpm{0.14}{0.01}}
& \tbp{\vpm{27.1}{0.8}}
& \tbp{\vpm{94.7}{1.7}}
& \tbp{\vpm{96.8}{1.1}}
& \tbp{\vpm{0.16}{0.02}}
& \tbp{\vpm{29.2}{1.1}} \\

\bottomrule
\end{tabular}
\end{adjustbox}
\arrayrulecolor{black}
}%
\end{minipage}
\hfill
\begin{minipage}[t]{0.26\textwidth}
\centering
\caption{
Budget efficiency (FalconGym). $K_{90}$, $K_{95}^{\mathrm{wp}}$ are to reach $90\%$ success, $95\%$ waypoint following.
}
\label{tab:budget_efficiency}
{%
\scriptsize
\setlength{\tabcolsep}{2.6pt}
\renewcommand{\arraystretch}{1.14}
\arrayrulecolor{tablerule}
\begin{adjustbox}{max width=\linewidth,center}
\begin{tabular}{@{}lccc@{}}
\toprule
\textbf{Method}
& \makecell{\textbf{$K_{90}$}\\\metricdown}
& \makecell{\textbf{$K_{95}^{\mathrm{wp}}$}\\\metricdown}
& \makecell{\textbf{Success}\\\textbf{at $K=6$} (\%) \metricup} \\
\midrule

Rand
& $>12$
& $>12$
& \vpm{14.8}{2.9} \\

PK
& $10$
& $11$
& \vpm{68.3}{3.9} \\

PD
& $10$
& $11$
& \vpm{91.4}{2.1} \\

VG
& $7$
& $8$
& \vpm{76.9}{3.5} \\

DG
& $5$
& $6$
& \vpm{84.6}{3.0} \\

RO
& $3$
& $3$
& \vpm{98.2}{0.9} \\

\rowcolor{scodabg}
\tbp{SCoDA}
& \tbp{$3$}
& \tbp{$3$}
& \tbp{\vpm{96.5}{1.3}} \\

\bottomrule
\end{tabular}
\end{adjustbox}
\arrayrulecolor{black}
}%
\end{minipage}

\end{table}

\paragraph{Key Finding \#2: \tbp{SCoDA} generates
context-aware placements.}
\begin{wrapfigure}[17]{r}{0.5\textwidth}
    \centering
    \includegraphics[width=\linewidth]{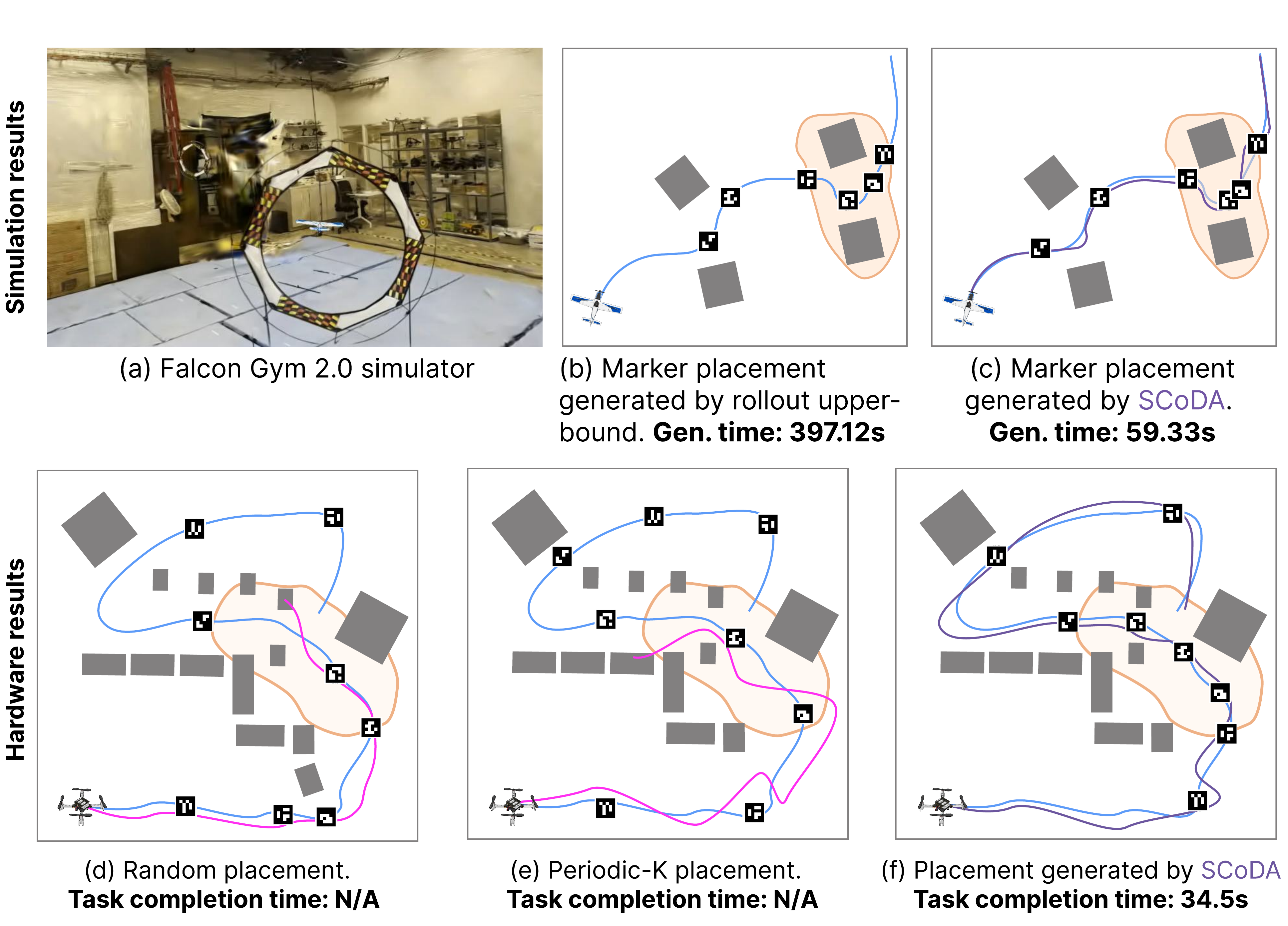}
    \caption{\tbp{SCoDA} generates context-aware placements on both simulated (top) and hardware (bottom) tasks, matching rollout-optimized quality orders of magnitude faster while baselines fail to complete the task.}
    \label{fig:qual}
\end{wrapfigure}
Figures~\ref{fig:placement_examples} and~\ref{fig:qual} show that \tbp{SCoDA}'s generated placements concentrate near disturbance entry points, obstacle-dense segments, and parts of the trajectory where tracking errors are likely to compound. This matters because each fiducial consumes one of only $K$ opportunities to correct the robot's pose estimate. Placing a fiducial in an easy segment, where the robot would remain well-localized anyway, wastes augmentation capacity that could instead be used near a failure-prone region. In contrast, \tbp{SCoDA} learns to allocate observations where they change the closed-loop execution outcome: before uncertainty grows too large, during regions where disturbances affect tracking, or shortly after those regions to help the robot recover. \tbp{SCoDA} improves reliability not by making the entire environment more observable, but by placing limited corrective observations where they are most useful for executing the planned trajectory.

\paragraph{Key Finding \#3: \tbp{SCoDA} approaches rollout-optimized placement without test-time search.}
\begin{wraptable}{r}{0.5\textwidth}
\centering
\caption{
Placement-time compute. Averages are over held-out  tasks; $K=6$.
}
\label{tab:compute}
{%
\setlength{\tabcolsep}{2.5pt}
\renewcommand{\arraystretch}{1.14}
\arrayrulecolor{tablerule}
\begin{adjustbox}{max width=\linewidth,center}
\begin{tabular}{@{}lcccc@{}}
\toprule
\textbf{Method}
& \makecell{\textbf{\# Rollouts}\\}
& \makecell{\textbf{Placement}\\\textbf{time} (s) \metricdown}
& \makecell{\textbf{Success}\\\textbf{rate} (\%) \metricup}
& \makecell{\textbf{Completion}\\\textbf{time} (s) \metricdown} \\
\midrule

PK
& $0$
& \vpm{0.04}{0.01}
& \vpm{68.3}{3.9}
& \vpm{79.1}{3.4} \\

VG
& $0$
& \vpm{0.47}{0.08}
& \vpm{76.9}{3.5}
& \vpm{71.8}{3.0} \\

DG
& $20$
& \vpm{3.86}{0.54}
& \vpm{84.6}{3.0}
& \vpm{63.7}{2.7} \\

RO
& $4500$
& \vpm{417.3}{34.6}
& \vpm{98.2}{0.9}
& \vpm{52.7}{1.4} \\

\rowcolor{scodabg}
\tbp{SCoDA}
& \tbp{$0$}
& \tbp{\vpm{0.18}{0.03}}
& \tbp{\vpm{94.8}{1.8}}
& \tbp{\vpm{56.1}{1.9}} \\


\bottomrule
\end{tabular}
\end{adjustbox}
\arrayrulecolor{black}
}%
\vspace{-1 em}
\end{wraptable}
Rollout-Opt has access to closed-loop evaluations for each test task and directly optimizes Eq.~\eqref{eq:placement_problem}, making it an expensive upper bound. \tbp{SCoDA} comes close to this upper bound using only a learned generator and shielded inference (refer Tables~\ref{tab:budget_efficiency} and~\ref{tab:compute} and Figure~\ref{fig:qual}). The remaining gap is the cost of amortization; the gain is that placement no longer requires expensive rollout search.

\paragraph{Key Finding \#4: amortization removes the placement-time bottleneck.}
Table~\ref{tab:compute} separates placement quality from placement cost. Rollout-Opt achieves excellent execution by evaluating 
many candidate layouts, while fast methods rely on surrogate objectives that leave performance on the table. \tbp{SCoDA} gives a better accuracy-compute trade-off: a single sample is already competitive, and shield sampling improves performance while remaining far cheaper than rollout optimization.

\paragraph{Ablations.}
\begin{wraptable}[12]{r}{0.50\textwidth}
\centering
\caption{
Ablations on disturbed simulation tasks with $K=6$ over 150 rollouts.
}
\label{tab:ablations}
{%
\scriptsize
\setlength{\tabcolsep}{2.6pt}
\renewcommand{\arraystretch}{1.14}
\arrayrulecolor{tablerule}
\begin{adjustbox}{max width=\linewidth,center}
\begin{tabular}{@{}lcccc@{}}
\toprule
\textbf{Variant}
& \makecell{\textbf{Success}\\(\%) \metricup}
& \makecell{\textbf{Waypoint}\\\textbf{follow.} (\%) \metricup}
& \makecell{\textbf{Completion}\\\textbf{time} (s) \metricdown}
& \makecell{\textbf{$J_{\mathrm{task}}$}\\\metricdown} \\
\midrule

\rowcolor{scodabg}
\tbp{SCoDA} full
& \tbp{\vpm{94.7}{1.7}}
& \tbp{\vpm{96.8}{1.1}}
& \tbp{\vpm{58.3}{2.1}}
& \tbp{\vpm{0.18}{0.03}} \\

\textbf{\textcolor{red}{(-)}}  shield
& \vpm{78.1}{3.5}
& \vpm{85.3}{2.9}
& \vpm{75.9}{3.5}
& \vpm{0.74}{0.07} \\

\textbf{\textcolor{red}{(-)}}  interval term
& \vpm{82.4}{3.2}
& \vpm{88.1}{2.6}
& \vpm{70.8}{3.2}
& \vpm{0.62}{0.06} \\

\textbf{\textcolor{red}{(-)}}  separation term
& \vpm{86.2}{2.9}
& \vpm{90.7}{2.4}
& \vpm{66.6}{2.9}
& \vpm{0.51}{0.05} \\

\textbf{\textcolor{red}{(-)}}  rollout $\mathbf{y}$
& \vpm{84.9}{3.0}
& \vpm{89.4}{2.5}
& \vpm{68.2}{3.0}
& \vpm{0.43}{0.05} \\

\textbf{\textcolor{red}{(-)}}  disturb. cont. $\mathbf{d}$
& \vpm{80.7}{3.4}
& \vpm{85.9}{3.0}
& \vpm{74.5}{3.4}
& \vpm{0.58}{0.06} \\

\textbf{\textcolor{red}{(-)}} CFG
& \vpm{88.5}{2.6}
& \vpm{92.1}{2.0}
& \vpm{62.9}{2.6}
& \vpm{0.36}{0.04} \\

post-hoc filtering only
& \vpm{87.6}{2.8}
& \vpm{91.6}{2.2}
& \vpm{64.1}{2.8}
& \vpm{0.28}{0.04} \\

workspace diffusion
& \vpm{70.4}{3.9}
& \vpm{78.8}{3.5}
& \vpm{83.7}{3.8}
& \vpm{0.91}{0.08} \\

\bottomrule
\end{tabular}
\end{adjustbox}
\arrayrulecolor{black}
}%
\vspace{-2.0 em}
\end{wraptable}
Table~\ref{tab:ablations} identifies the main sources of improvement. Removing the shield reduces critical-region support and increases redundant placements; the interval and separation terms direct corrections to important regions and prevent budget clustering. Removing the rollout profile weakens the link between generated layouts and execution quality, while removing disturbance context reduces task-specific responsiveness. Post-hoc filtering improves feasibility less than shielded denoising since it selects after generation rather than shaping samples during it. Workspace-coordinate diffusion performs worst among learned variants, confirming that trajectory-indexed generation is central to sparse augmentation.


\section{Conclusion and Limitations}
\label{sec:conclusion}

\tbp{SCoDA} demonstrates that diffusion can be used as a generative designer for just-enough environment augmentation. The method couples a trajectory-indexed augmentation representation, rollout-conditioned diffusion over sparse fiducial layouts, and a task-space shield that focuses the finite budget on useful corrective observations. Across simulation and hardware, \tbp{SCoDA} (i) improves trajectory-execution reliability over strong heuristic and learned baselines, (ii) reaches $90\%$ success and $95\%$ waypoint following using the same minimum number of fiducials as the rollout-optimized upper bound, and (iii) produces shield-ranked layouts in under a second rather than requiring hundreds of seconds of closed-loop search. The results show that reliable execution under partial observability improves when task-aware fiducials are placed where localization support is most useful.

\textbf{Limitations:}
\tbp{SCoDA} optimizes where to place a prescribed number $K$ of markers, but it does not choose $K$. \tbp{SCoDA} assumes that a reference trajectory and mapped environment are available before deployment, so it is best suited to structured tasks. Because training supervision comes from closed-loop rollouts, performance depends on how well the models and estimator match the deployment setting. 
\textbf{Future work:} Our experiments focus on single-robot trajectory execution. Extending \tbp{SCoDA} to multi-robot settings, where cooperative agents can act as moving sources of localization information or share relative observations, is a promising direction. Another important extension includes jointly choosing the number and placement of fiducials and reasoning about richer three-dimensional visibility.
\newpage
\bibliography{corl_2026_template_submission/references, corl_2026_template_submission/example}  

\newpage

\appendix
\begin{appendix}

\clearpage
\startcontents[appendix]

\section*{\normalsize Appendix Contents}

\begingroup
\footnotesize
\printcontents[appendix]{}{1}{\setcounter{tocdepth}{2}}
\endgroup

\clearpage

\section{Experiment setting and Details} \label{expt}
\subsection{Simulation Setup: FalconGym}
\label{app:falcongym_setup}

\begin{figure}[H]
    \centering
    \includegraphics[width=1\linewidth]{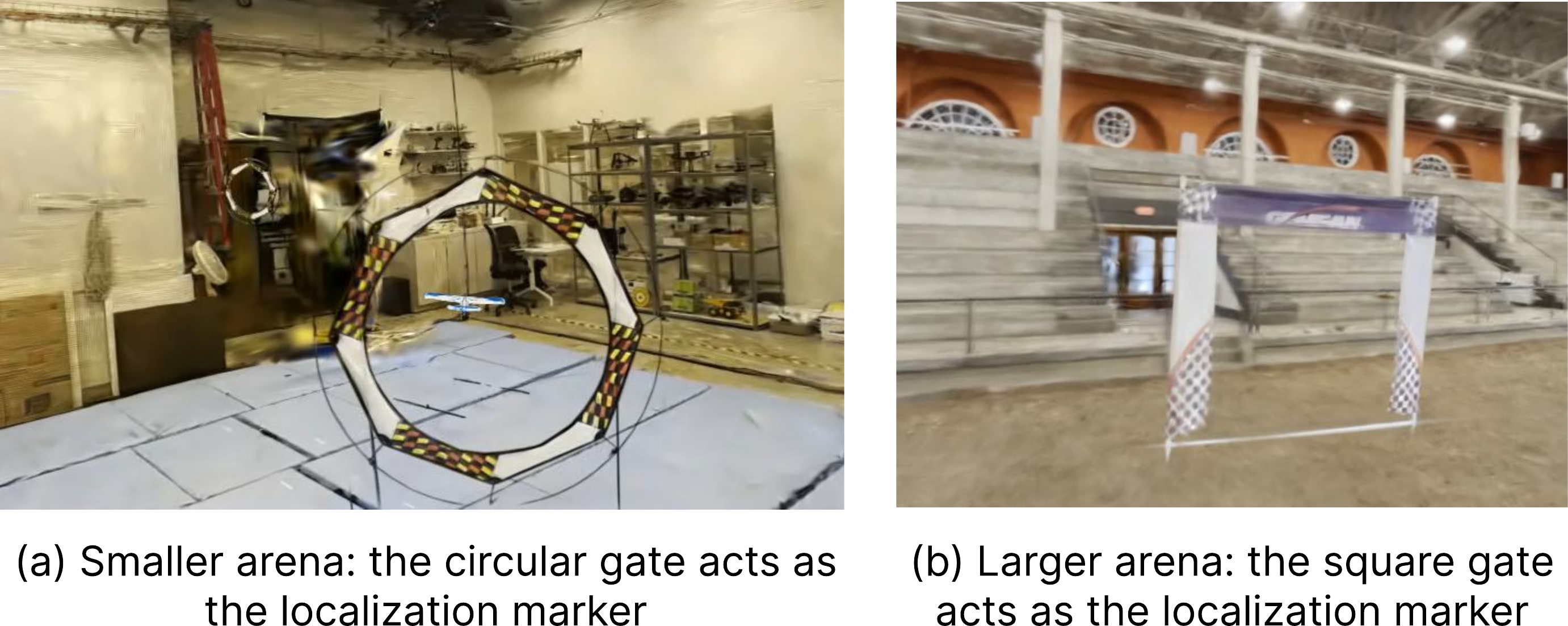}
    \caption{Simulator environment}
    \label{fig:falcongym-img}
\end{figure}

We evaluate \tbp{SCoDA} in FalconGym 2.0~\cite{miao2026performanceguided}, using the fixed-wing vehicle dynamics from FalconWing~\cite{miao2025falconwing}. FalconGym provides a photorealistic aerial-navigation simulator with editable scene elements, which allows us to insert task-specific fiducial structures while keeping the vehicle dynamics, reference trajectory, estimator, controller, and rendering pipeline fixed across methods. We use the simulator only as an execution environment; \tbp{SCoDA} does not receive closed-loop rollout evaluations at test time.

Figure~\ref{fig:falcongym-img} shows the simulated fiducial structures used in our FalconGym experiments. We modify FalconGym so that these fiducial structures act as localization-update regions. During execution, when the fixed-wing vehicle reaches a pose from which an inserted fiducial is detectable under the nominal camera and visibility model, the estimator receives a noisy pose correction associated with the known fiducial pose. This models the effect of visual landmarks such as AprilTags or gate-like fiducials without changing the low-level dynamics or controller. The pose-correction noise, detection range, field-of-view threshold, and visibility checks are shared across all augmentation methods.

We also extend the simulator with localized wind regions for the disturbance-augmented tasks. Each wind region occupies a specified spatial region of the environment and applies an external disturbance to the vehicle while it is inside that region. The disturbance interval used by \tbp{SCoDA}'s context is obtained by projecting the wind-affected segment of the nominal trajectory into trajectory-progress coordinates. Nominal tasks use the same simulator, estimator, controller, and trajectory-generation procedure, but do not include external wind disturbances.

We evaluate on two arena sizes: a small arena of size $5.53\,\mathrm{m}\times3.64\,\mathrm{m}$ and a larger room of size $10.71\,\mathrm{m}\times8.92\,\mathrm{m}$. The main simulated execution table averages results over both room sizes. The budget-efficiency and placement-time compute tables are reported on the larger room only, since it provides a wider range of feasible sparse-placement configurations and larger observation-poor regions. All simulation evaluations are performed on held-out settings not used to train \tbp{SCoDA}; the held-out split includes unseen trajectories, unseen goals, unseen disturbance placements and unseen room layouts. Unless otherwise stated, simulated fixed-budget experiments use $K=6$ fiducials and report means and standard errors over $150$ stochastic closed-loop rollouts.



\begin{table}[h]
\centering
\caption{FalconGym simulation parameters}
\label{tab:falcongym_setup}
\scriptsize
\setlength{\tabcolsep}{4pt}
\renewcommand{\arraystretch}{1.12}

\begin{tabular}{@{}l!{\vrule width 0.1pt}ll@{}}
\toprule
{\thcv{Parameter}} & {\thcv{Meaning}} & {\thcv{Value}} \\
\midrule
\cellcolor{scodaVioletTint}\emph{Environment setup} & & \\
Small arena size & Planar room dimensions & $5.53\,\mathrm{m}\times3.64\,\mathrm{m}$ \\
Large room size & Planar room dimensions & $10.71\,\mathrm{m}\times8.92\,\mathrm{m}$ \\
Simulator & Photorealistic aerial-navigation environment & FalconGym 2.0~\cite{miao2026performanceguided} \\
Vehicle model & Fixed-wing dynamics used in simulation & FalconWing~\cite{miao2025falconwing} \\

\midrule
\cellcolor{scodaVioletTint}\emph{Evaluation setup} & & \\
Fixed-budget fiducials & Number of added fiducials in main simulation experiments & $K=6$ \\
Rollouts per reported setting & Stochastic closed-loop executions used for evaluation & $150$ \\
Train/test split & Held-out evaluation settings & 80/20 \\

\midrule
\cellcolor{scodaVioletTint}\emph{Planning and sensing setup} & & \\
Reference-trajectory generation & Planner or trajectory source & Oracle path planner (A*) \\
Detection range & Minimum and maximum fiducial-detection range & $14\,\mathrm{cm}$ \\
Camera field of view & Visibility threshold used for fiducial detection & $15\,\mathrm{cm}$ \\
Number of wind regions & Wind regions per disturbed task & 1 \\
\bottomrule
\end{tabular}
\end{table}

\subsection{Hardware Setup: Crazyflie Deployment}
\label{app:hardware_setup}

\begin{figure}[H]
    \centering
    \includegraphics[width=1\linewidth]{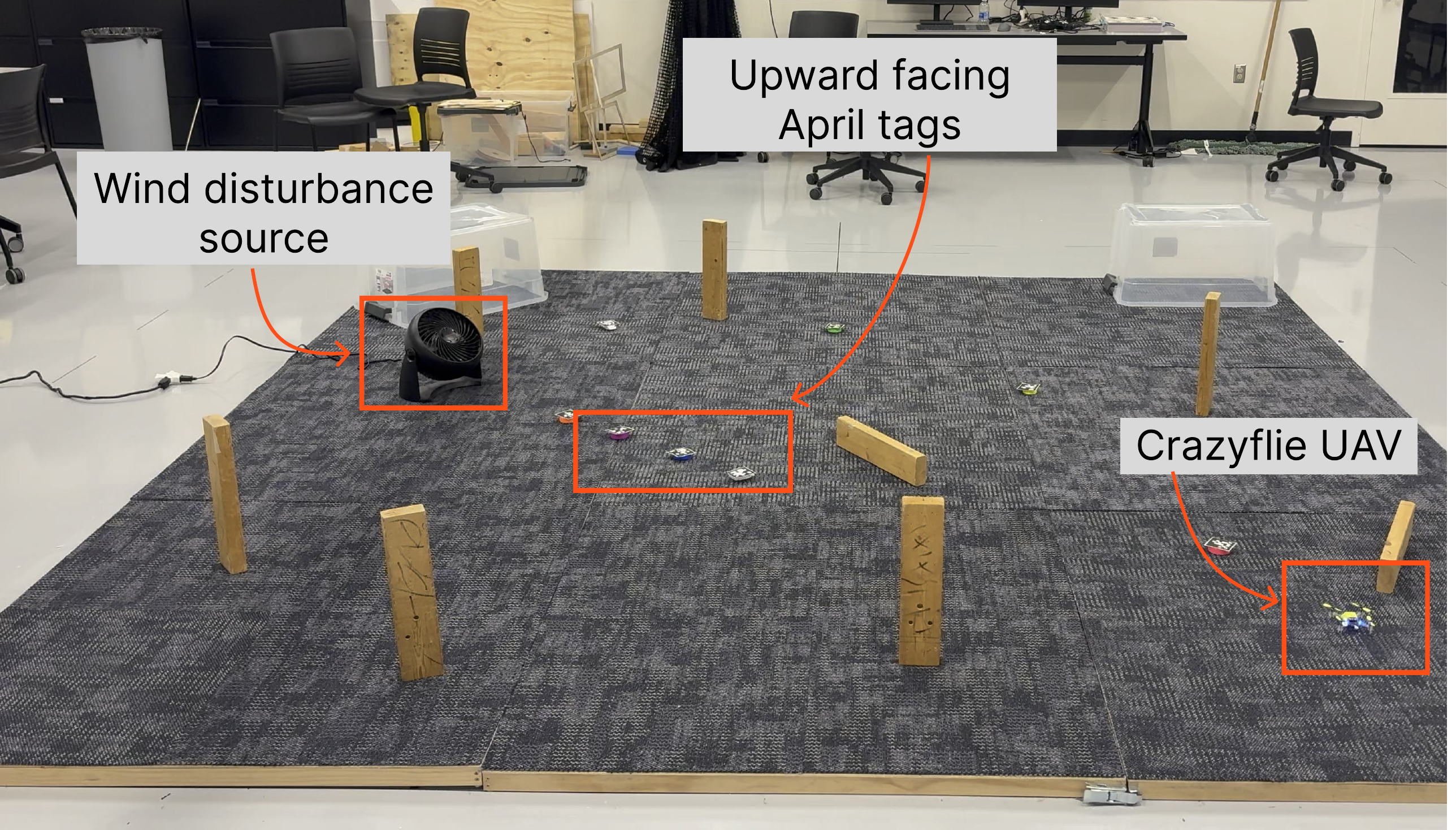}
    \caption{Hardware environment setup}
    \label{fig:hardware-img}
\end{figure}

We also evaluate \tbp{SCoDA} on a Crazyflie quadrotor in a $2\,\mathrm{m}\times2\,\mathrm{m}$ indoor arena. The hardware experiments are designed to test whether the trajectory-progress placements generated by \tbp{SCoDA} provide useful localization support during real closed-loop execution. The estimator, low-level flight controller, reference trajectory, and fiducial-detection pipeline are kept fixed across methods.

Figure~\ref{fig:hardware-img} shows the hardware deployment. At deployment time, \tbp{SCoDA} generates a sparse set of desired correction locations along the planned trajectory. We instantiate these suggestions using physical AprilTags placed on the floor of the arena. The tag poses are measured in the arena coordinate frame and provided to the estimator before flight. During execution, when the Crazyflie detects a tag, the known tag pose and the measured camera-to-tag transform provide a pose correction to the robot estimator. This follows the standard fiducial-localization setup in which tags have known identities and calibrated world-frame poses.

The hardware study is intended primarily to evaluate the quality of the generated correction schedule and sparse placement pattern, not the difficulty of finding arbitrary wall or ceiling mounting surfaces. Therefore, the physical-placement problem is controlled: tags are placed on the floor near the generated trajectory-progress locations and oriented so that a robot following the nominal trajectory, or deviating moderately from it, can still observe the tags. This makes the hardware experiment a direct test of whether \tbp{SCoDA} allocates a limited number of corrections to useful parts of the task trajectory. All methods use the same floor-placement convention, the same tag-size and detection settings, and the same calibrated tag-pose map.

The hardware arena contains obstacles, and the obstacle placements used for evaluation are held out from training. The evaluated trajectories are also previously unseen. Thus, the hardware results test deployment on unseen obstacle configurations and unseen reference trajectories rather than memorization of a training layout. Unless otherwise stated, hardware experiments use $K=8$ AprilTags and report results over 45 total trials.

\begin{table}[h]
\centering
\caption{Hardware deployment parameters.}
\label{tab:hardware_setup}
\scriptsize
\setlength{\tabcolsep}{4pt}
\renewcommand{\arraystretch}{1.12}

\begin{tabular}{@{}l!{\vrule width 0.1pt}ll@{}}
\toprule
{\thcv{Parameter}} & {\thcv{Meaning}} & {\thcv{Value}} \\
\midrule
Robot platform & Quadrotor used for hardware deployment & Crazyflie 2.0 \\
Arena size & Planar hardware workspace dimensions & $2\,\mathrm{m}\times2\,\mathrm{m}$ \\
Fiducial type & Physical visual markers used for localization & AprilTags \\
Fiducial budget & Number of deployed tags & $K=8$ \\
Tag placement surface & Physical mounting surface & Floor \\
Controller & Low-level and/or trajectory-tracking controller & Cascaded PID \\
Waypoint-following criterion & Definition of hardware waypoint-following metric & The drone moves within a radius of 13cm of the given waypoint \\
Obstacle configurations & Held-out obstacle-placement protocol & Previously unseen \\
Number of trials & Hardware trials used in reported table & 45 \\
\bottomrule
\end{tabular}
\end{table}


\subsection{Baselines}
We compare \tbp{SCoDA} against baselines chosen to isolate trajectory-indexed generation, rollout-profile conditioning, shielded denoising, and test-time rollout optimization. For all methods, the reference trajectory $\tau$, mapped environment $\mathcal{E}$, estimator, controller, camera model, process-noise model, measurement-noise model, and disturbance model are held fixed. For all fixed-budget augmentation methods, the fiducial budget $K$ and physical feasibility constraints are also held fixed. Let $\mathcal{P}$ denote the discretized feasible physical candidate set used by physical-space baselines. A pose is included in $\mathcal{P}$ only if it satisfies the common installability, range, field-of-view, marker-orientation, occlusion, and workspace-duplicate constraints. Unless otherwise stated, fixed-budget augmentation methods return $\mathbf{m}\in\mathcal{M}_K$, while Periodic-Dense returns $\mathbf{m}_{\mathrm{PD}}\in\mathcal{M}_{K_{\mathrm{PD}}}$. Methods that choose desired correction locations in trajectory-progress space are mapped to physical fiducial poses using the same instantiation map $M_{\mathcal{E},\tau}$. If $M_{\mathcal{E},\tau}$ cannot instantiate a requested progress value, the value is snapped to the nearest unused feasible progress value within the allowed local progress window; if no such value exists, the layout is rejected and the method continues with its next candidate or proposal.

\textit{No-Augmentation \textbf{(NA)}} uses no added fiducials. The robot executes the reference trajectory using the same controller, estimator, process noise, measurement noise, and disturbance model as all other methods, but receives no additional fiducial-based pose corrections. This baseline measures how often the planned trajectory can be executed using the unaugmented environment alone.

\textit{Random \textbf{(Rand)}} samples $K$ fiducial poses uniformly without replacement from the feasible physical candidate set $\mathcal{P}$. Rand therefore tests whether reliability improves merely from adding feasible visual structure, without using trajectory timing, disturbance context, localizability, predicted deviation, or rollout outcomes.

\textit{Periodic-$K$ \textbf{(PK)}} places $K$ fiducials at uniformly spaced progress values along the reference trajectory and then instantiates them physically with $M_{\mathcal{E},\tau}$. The target progress values are the centers of $K$ equal progress bins,
\begin{equation}
v_i^{\mathrm{PK}}=\frac{i-\frac{1}{2}}{K},\qquad i=1,\ldots,K.
\end{equation}
PK is a strong non-learning baseline because it distributes corrections across the full execution, but it is not task-aware: it spends the same budget in easy, already-localizable regions as in disturbance, recovery, or terminal-approach regions.

\textit{Periodic-Dense \textbf{(PD)}} is an unbounded periodic-coverage baseline. Instead of using a fixed budget, PD increases the periodic budget $K_{\mathrm{PD}}$ until the maximum nominal observation gap along the reference trajectory is below a threshold $\Delta_{\mathrm{obs}}$. Let $o_n(\mathbf{m})=\mathds{1}[\text{at least one added fiducial in }\mathbf{m}\text{ is detectable from }x_n^r]$. In implementation, the observation gap is computed on the discretized reference trajectory as
\begin{equation}
\mathcal{G}(\mathbf{m})=\frac{1}{N}\max\{b-a+1:o_n(\mathbf{m})=0\text{ for all }n=a,\ldots,b\}.
\end{equation}
PD chooses the smallest periodic budget $K_{\mathrm{PD}}$ such that
\begin{equation}
\mathcal{G}(\mathbf{m}_{\mathrm{PD}})\leq\Delta_{\mathrm{obs}}.
\end{equation}
PD is not a fair fixed-budget competitor to \tbp{SCoDA}; it measures how many markers are needed when one solves the problem by dense periodic coverage rather than task-aware sparse augmentation.

\textit{Critical-Region \textbf{(CR)}} is a hand-designed disturbance-aware heuristic with access to the same disturbance and recovery intervals used by \tbp{SCoDA}'s shield. For disturbance-conditioned tasks, CR allocates fiducials to the pre-disturbance, disturbance-traversal, recovery, and terminal-approach regions, then maps the selected progress values to physical marker poses using $M_{\mathcal{E},\tau}$. When the budget is smaller than the number of critical regions, CR assigns markers according to the fixed priority order pre-disturbance, disturbance traversal, recovery, and terminal approach. When extra markers remain, they are distributed across the critical regions in proportion to interval length; within each assigned interval, progress values are placed at equally spaced bin centers. CR is omitted from nominal settings because its allocation rule is defined by disturbance and recovery intervals.

\textit{Visibility-Greedy \textbf{(VG)}} greedily selects fiducials that make the largest number of nominal reference states observable. For each candidate marker pose $p\in\mathcal{P}$, let $\mathcal{O}(p)\subseteq\{0,\ldots,N\}$ be the set of reference-trajectory indices from which $p$ is detectable under the same camera, projected-size, viewing-geometry, and occlusion model used by $M_{\mathcal{E},\tau}$. Starting from $\mathcal{S}_0=\emptyset$, VG repeatedly selects
\begin{equation}
p_j\in\arg\max_{p\in\mathcal{P}\setminus\mathcal{S}_{j-1}}\left|\mathcal{O}(p)\setminus\bigcup_{q\in\mathcal{S}_{j-1}}\mathcal{O}(q)\right|,
\end{equation}
and sets $\mathcal{S}_j=\mathcal{S}_{j-1}\cup\{p_j\}$ until $K$ markers are selected. Ties are broken by the common candidate ordering. VG uses geometry and visibility but ignores whether the covered states are important for closed-loop execution.

\textit{Localizability-Greedy \textbf{(LG)}} greedily selects fiducials using a geometric localizability surrogate inspired by camera-localizability marker-placement objectives~\cite{huang2023optimizing}, rather than a binary visibility count. For each feasible candidate marker pose $p$ and reference index $n$, let $\chi(n,p)$ indicate whether $p$ is detectable from $x_n^r$, let $r(n,p)$ be the camera-marker range, let $\theta(n,p)$ be the marker incidence angle, and let $A(n,p)$ be the projected marker area. We use the normalized score
\begin{equation}
\ell(n,p)=\chi(n,p)\left(1-\frac{r(n,p)-r_{\min}}{r_{\max}-r_{\min}}\right)\max(0,\cos\theta(n,p))\min\left(1,\frac{A(n,p)}{A_{\mathrm{ref}}}\right),
\end{equation}
where $A_{\mathrm{ref}}=\max_{n,p:\chi(n,p)=1}A(n,p)$. LG maintains the best localizability score already provided at each reference state and greedily selects the candidate with the largest marginal improvement,
\begin{equation}
p_j\in\arg\max_{p\in\mathcal{P}\setminus\mathcal{S}_{j-1}}\sum_{n=0}^{N}\left[\max\!\left(L_{j-1}(n),\ell(n,p)\right)-L_{j-1}(n)\right],
\end{equation}
where $L_{j-1}(n)=\max_{q\in\mathcal{S}_{j-1}}\ell(n,q)$ and $L_0(n)=0$. Ties are broken by the common candidate ordering. LG is stronger than VG because it accounts for nominal measurement quality, but it still optimizes a geometric-localization surrogate rather than closed-loop task success.

\textit{Deviation-Greedy \textbf{(DG)}} is a deviation-inspired heuristic related to landmark-placement methods that reason about predicted deviation from a desired trajectory~\cite{beinhofer2013effective}. DG places fiducials near trajectory regions where the robot is predicted to accumulate large tracking error without additional localization support. DG first estimates a deviation profile using $B_{\mathrm{DG}}=20$ stochastic unaugmented rollouts. Let $\tilde{e}_{n,b}^{\tau}$ denote the tracking error at reference index $n$ in rollout $b$ if that index is reached before termination, and let $\|\tilde{e}_{n,b}^{\tau}\|_Q^2=E_{\max}$ if the rollout terminates before reaching index $n$. The risk score is
\begin{equation}
r(n)=\frac{1}{B_{\mathrm{DG}}}\sum_{b=1}^{B_{\mathrm{DG}}}\min\left(\|\tilde{e}_{n,b}^{\tau}\|_Q^2,E_{\max}\right).
\end{equation}
DG sorts reference indices by decreasing $r(n)$ and greedily selects the highest-risk index whose normalized progress is at least $\delta_{\mathrm{DG}}$ away from all previously selected indices, until $K$ indices are selected. The selected progress values are then instantiated with $M_{\mathcal{E},\tau}$. Unlike RO, DG does not evaluate candidate fiducial layouts under closed-loop execution; it only estimates where the unaugmented trajectory is likely to drift.

\textit{Rollout-Opt \textbf{(RO)}} is an expensive test-time rollout-search baseline that directly evaluates candidate layouts using the closed-loop objective in Eq.~\eqref{eq:placement_problem}. RO performs random shooting over progress-space layouts: it samples $K$ progress values from the feasible progress grid subject to the same minimum-separation rule used by the validation check, instantiates the resulting layout with $M_{\mathcal{E},\tau}$, rejects physically infeasible layouts, and evaluates each retained candidate using $B_{\mathrm{RO}}$ stochastic rollouts. For each candidate layout $\mathbf{m}$, RO estimates
\begin{equation}
\widehat{J}_{\mathrm{RO}}(\mathbf{m})=\frac{1}{B_{\mathrm{RO}}}\sum_{b=1}^{B_{\mathrm{RO}}}\left[\sum_{t=0}^{T_{\max}}\|e_{t,b}^{\tau}\|_Q^2+\lambda_{\mathrm{time}}t_{f,b}+C_{\mathrm{fail}}\mathds{1}[\mathcal{F}_{\tau,b}]\right].
\end{equation}
RO returns the lowest-cost layout found within its total rollout budget. In the compute comparison, this total budget is $4500$ closed-loop rollouts. RO has access to closed-loop evaluations for each test task, so it is not an amortized placement method and is much more expensive than \tbp{SCoDA}. We use it as an empirical test-time optimization reference rather than as a formal global optimum.

We also report ablations of \tbp{SCoDA} to isolate the contribution of each component. Removing the shield tests whether rollout-conditioned diffusion alone uses the finite marker budget effectively. Removing the interval term tests whether explicitly covering task-relevant regions is necessary. Removing the separation term tests whether progress-space redundancy hurts sparse augmentation. Removing the rollout profile $\mathbf{y}$ tests the value of conditioning generation on execution outcomes. Removing disturbance context $\mathbf{d}$ tests whether \tbp{SCoDA} adapts to task-specific perturbations. Removing classifier-free guidance tests the effect of stronger conditioning at inference. Post-hoc filtering samples from the unshielded diffusion model and then applies the same final hard feasibility check, testing whether rejecting bad samples after generation is sufficient compared with shaping the reverse process itself. Workspace diffusion replaces trajectory-indexed generation with direct workspace-coordinate generation before applying the same physical feasibility checks, testing whether reasoning in trajectory-progress space is important for sparse task-aware augmentation.

\section{Derivations and Implementation Parameters}
\label{app:shield}

This appendix gives the derivations and implementation details referenced in the main paper. 

\subsection{Notation}

In this appendix, $t$ indexes the diffusion step, not physical rollout time, and $\mathbf{x}_t$ denotes the noisy diffusion variable rather than the robot state. Clean training augmentations are represented as vectors $\mathbf{x}_0\in[-1,1]^K$, where $K$ is the fiducial budget. During sampling, the reverse process produces an unconstrained denoised estimate, which is clipped to $[-1,1]^K$ before being interpreted as a progress-space augmentation. We convert diffusion coordinates to normalized trajectory-progress coordinates by
\begin{equation}
S(\mathbf{u})=\frac{\mathbf{u}+\mathbf{1}}{2},\qquad \mathbf{v}=S(\mathbf{u})\in[0,1]^K.
\end{equation}

Let $\operatorname{sg}[\cdot]$ denote the stop-gradient operator. To avoid zeroing the shield gradient when a predicted clean value lies outside $[-1,1]$, we use a straight-through clip for shield evaluation:
\begin{equation}
\operatorname{clip}_{\mathrm{ST}}(z)=z+\operatorname{sg}\!\left[\operatorname{clip}_{[-1,1]}(z)-z\right].
\end{equation}
The forward value of $\operatorname{clip}_{\mathrm{ST}}$ equals the clipped value, while its backward derivative with respect to $z$ is the identity. For stable and inexpensive guidance, we also stop gradients through the denoiser prediction when computing the shield gradient. The predicted clean sample used by the shield is
\begin{equation}
\widehat{\mathbf{x}}_{0,t}^{\mathrm{sg}}=\frac{\mathbf{x}_t-\sqrt{1-\bar{\alpha}_t}\operatorname{sg}\!\left[\hat{\epsilon}_{\mathrm{cfg}}(\mathbf{x}_t,t,\mathbf{c})\right]}{\sqrt{\bar{\alpha}_t}},
\end{equation}
and the corresponding progress vector is
\begin{equation}
\mathbf{v}_t=S\!\left(\operatorname{clip}_{\mathrm{ST}}\!\left(\widehat{\mathbf{x}}_{0,t}^{\mathrm{sg}}\right)\right).
\end{equation}
The denoising network parameters are frozen during sampling; only the derivative with respect to the current diffusion state $\mathbf{x}_t$ is used for shielding.

\subsection{Task-Relevant Intervals}

The context $\mathbf{c}$ specifies task-relevant trajectory intervals $\mathcal{I}(\mathbf{c})=\{[a_m,b_m]\}_{m=1}^{M}$, where $0\leq a_m\leq b_m\leq1$. In disturbance-conditioned tasks, the disturbance context is $\mathbf{d}=[u_{d,s},u_{d,e},u_{r,e}]$, where $u_{d,s}$ is the disturbance-entry progress, $u_{d,e}$ is the disturbance-exit progress, and $u_{r,e}$ is the end of the recovery region. Unless otherwise stated, we construct
\begin{equation}
\mathcal{I}(\mathbf{c})=\left\{[\max(0,u_{d,s}-\Delta_{\mathrm{pre}}),u_{d,s}],\ [u_{d,s},u_{d,e}],\ [u_{d,e},u_{r,e}],\ [\max(0,1-\Delta_{\mathrm{goal}}),1]\right\}.
\end{equation}
The four intervals correspond to pre-disturbance correction, disturbance traversal, recovery, and terminal approach. Intervals with width smaller than $\epsilon_{\mathrm{int}}$ are omitted. For nominal tasks without an external disturbance interval, we use the terminal-approach interval
\begin{equation}
\mathcal{I}_{\mathrm{nom}}(\mathbf{c})=\left\{[\max(0,1-\Delta_{\mathrm{goal}}),1]\right\}.
\end{equation}
When intervals overlap, a single progress value may contribute to multiple interval counts. This is intentional because overlapping task-relevant regions can be supported by the same fiducial-induced correction.

For interval $m$, let $n_m$ be the required number of fiducials in that interval. In all reported experiments using the interval-coverage shield, $n_m=1$ unless otherwise stated. We use the sigmoid $\sigma(r)=1/(1+\exp(-r))$ and the smooth interval-membership score
\begin{equation}
\rho(v;a,b)=\sigma(\kappa(v-a))\sigma(\kappa(b-v)),\qquad \kappa>0.
\end{equation}
The score $\rho(v;a,b)$ is close to one when $v$ lies well inside an interval whose width is large relative to $1/\kappa$, and close to zero far outside the interval.

To avoid a nondifferentiability in the separation penalty, we use the smooth distance
\begin{equation}
d_{ij}^{\eta}=\sqrt{(v_i-v_j)^2+\eta^2},
\end{equation}
where $\eta>0$ is small. The shield objective is
\begin{equation}
\label{eq:app_j_task}
J_{\mathrm{task}}(\mathbf{v};\mathbf{c})=\sum_{m=1}^{M}\beta_m\left[n_m-\sum_{k=1}^{K}\rho(v_k;a_m,b_m)\right]_++\beta_s\sum_{i<j}\sigma\!\left(\kappa_s(\delta_{\min}-d_{ij}^{\eta})\right).
\end{equation}
The first term is a differentiable surrogate for requiring at least $n_m$ progress values inside interval $m$. The second term penalizes progress-space redundancy under a finite marker budget. The positive-part function $[r]_+=\max(r,0)$ is implemented using the automatic-differentiation subgradient at $r=0$. The separation term encodes the design assumption that, under a small marker budget, multiple fiducials observed at nearly the same trajectory progress are usually less valuable than corrections distributed across the execution. It is a progress-space redundancy heuristic, not a statement that simultaneous multi-tag observations are never useful.

\subsection{Exponential-Tilt View of Shielding}

Let $p_{\theta,\mathrm{base}}(\mathbf{u}\mid\mathbf{c})$ denote the distribution induced by the reverse sampler before applying the progress-space shield. In our experiments, this base sampler includes classifier-free guidance. For readability, we write this distribution as $p_\theta(\mathbf{u}\mid\mathbf{c})$. For a fixed context $\mathbf{c}$, define the tilted distribution over valid progress-space augmentations $\mathbf{u}\in[-1,1]^K$ by
\begin{equation}
\label{eq:app_tilt}
q_\lambda(\mathbf{u}\mid\mathbf{c})=\frac{p_\theta(\mathbf{u}\mid\mathbf{c})\exp\!\left(-\lambda J_{\mathrm{task}}(S(\mathbf{u});\mathbf{c})\right)}{Z_\lambda(\mathbf{c})},
\end{equation}
where $\lambda\geq0$ is the shielding strength and
\begin{equation}
Z_\lambda(\mathbf{c})=\int_{[-1,1]^K}p_\theta(\mathbf{u}\mid\mathbf{c})\exp\!\left(-\lambda J_{\mathrm{task}}(S(\mathbf{u});\mathbf{c})\right)d\mathbf{u}.
\end{equation}
Since $K$, $\beta_m$, and $\beta_s$ are finite and $J_{\mathrm{task}}$ is bounded on $[0,1]^K$, $Z_\lambda(\mathbf{c})$ is finite whenever $p_\theta$ is a normalized density on $[-1,1]^K$. For readability, write $J(\mathbf{u})=J_{\mathrm{task}}(S(\mathbf{u});\mathbf{c})$. Assuming differentiation under the integral is valid, we have
\begin{equation}
\mathbb{E}_{q_\lambda}[J]=\frac{\int_{[-1,1]^K}J(\mathbf{u})p_\theta(\mathbf{u}\mid\mathbf{c})\exp(-\lambda J(\mathbf{u}))d\mathbf{u}}{Z_\lambda(\mathbf{c})}=-\frac{d}{d\lambda}\log Z_\lambda(\mathbf{c}).
\end{equation}
Differentiating once more gives
\begin{equation}
\frac{d}{d\lambda}\mathbb{E}_{q_\lambda}[J]=-\frac{d^2}{d\lambda^2}\log Z_\lambda(\mathbf{c})=-\left(\mathbb{E}_{q_\lambda}[J^2]-\mathbb{E}_{q_\lambda}[J]^2\right)=-\mathrm{Var}_{q_\lambda}(J)\leq0.
\end{equation}
Therefore, under exact sampling from $q_\lambda$, increasing $\lambda$ cannot increase the expected shield objective. This result motivates the shielded sampler, but the actual reverse-diffusion sampler is approximate; the monotonicity statement above is not a guarantee that every generated sample has lower $J_{\mathrm{task}}$, nor is it a guarantee of closed-loop execution success.

\subsection{Shielded DDPM Update}

The DDPM reverse mean using a noise prediction $\hat{\epsilon}$ is
\begin{equation}
\label{eq:app_ddpm_mean}
\mu_\theta(\mathbf{x}_t,t,\mathbf{c})=\frac{1}{\sqrt{\alpha_t}}\left(\mathbf{x}_t-\frac{\beta_t}{\sqrt{1-\bar{\alpha}_t}}\hat{\epsilon}\right),
\end{equation}
where $\beta_t$ is the diffusion variance schedule, $\alpha_t=1-\beta_t$, and $\bar{\alpha}_t=\prod_{s=1}^{t}\alpha_s$. The shield gradient is
\begin{equation}
\mathbf{g}_t=\nabla_{\mathbf{x}_t}J_{\mathrm{task}}(\mathbf{v}_t;\mathbf{c}).
\end{equation}
If gradient clipping is used, we replace $\mathbf{g}_t$ by
\begin{equation}
\mathbf{g}_t\leftarrow \mathbf{g}_t\min\left(1,\frac{g_{\max}}{\|\mathbf{g}_t\|_2+\varepsilon_g}\right),
\end{equation}
where $g_{\max}>0$ is the maximum gradient norm and $\varepsilon_g>0$ prevents division by zero. The shielded sampler uses
\begin{equation}
\label{eq:app_shield_eps}
\hat{\epsilon}_{\mathrm{sh}}=\hat{\epsilon}_{\mathrm{cfg}}+\lambda_t\mathbf{g}_t,
\end{equation}
where $\lambda_t\geq0$ is the shield strength at reverse step $t$. Substituting Eq.~\eqref{eq:app_shield_eps} into Eq.~\eqref{eq:app_ddpm_mean} gives
\begin{equation}
\mu_{\mathrm{sh}}=\mu_{\mathrm{cfg}}-\frac{\beta_t\lambda_t}{\sqrt{\alpha_t}\sqrt{1-\bar{\alpha}_t}}\mathbf{g}_t.
\end{equation}
Equivalently, the shield induces a mean-space step size
\begin{equation}
\eta_t=\frac{\beta_t\lambda_t}{\sqrt{\alpha_t}\sqrt{1-\bar{\alpha}_t}},
\end{equation}
so that $\mu_{\mathrm{sh}}=\mu_{\mathrm{cfg}}-\eta_t\mathbf{g}_t$. Thus, adding $+\lambda_t\nabla_{\mathbf{x}_t}J_{\mathrm{task}}$ to the predicted noise moves the reverse mean in the negative gradient direction of $J_{\mathrm{task}}$. The positive sign in Eq.~\eqref{eq:app_shield_eps} is therefore consistent with descending the shield objective under the DDPM noise-prediction parameterization. In implementation, $\lambda_t$ is tuned with the timestep-dependent preconditioning above in mind. This calculation establishes the local direction of the reverse-mean shift induced by the $\epsilon$-space modification. It should not be interpreted as a proof that each stochastic reverse step decreases $J_{\mathrm{task}}$, because the sampler also includes diffusion noise, the denoiser score, clipping or projection, and a shield objective evaluated through a predicted clean sample.

Shielding is applied only on a specified subset of reverse steps $\mathcal{T}_{\mathrm{sh}}\subseteq\{1,\ldots,T\}$. Early reverse steps are often too noise-dominated for the progress-space objective to be meaningful, while very late steps leave little freedom to change the sample.

\subsection{Physical Instantiation and Final Feasibility Check}

After denoising, we obtain $\widehat{\mathbf{u}}=\operatorname{clip}_{[-1,1]}(\widehat{\mathbf{x}}_0)$ and $\widehat{\mathbf{v}}=S(\widehat{\mathbf{u}})$. The physical instantiation map returns either a feasible physical layout and realized progress values or a rejection:
\begin{equation}
(\widehat{\mathbf{m}},\widetilde{\mathbf{v}})=M_{\mathcal{E},\tau}(\widehat{\mathbf{u}}).
\end{equation}
Here $\widehat{\mathbf{m}}$ is the physical fiducial layout and $\widetilde{\mathbf{v}}\in[0,1]^K$ contains the realized correction progress values induced by the instantiated markers under the nominal reference trajectory. We use $\widetilde{\mathbf{v}}$, rather than the requested progress values $\widehat{\mathbf{v}}$, for the final hard feasibility check because $M_{\mathcal{E},\tau}$ may snap a requested progress value to a nearby feasible physical marker pose.

For each requested progress value $\widehat{v}_k$, $M_{\mathcal{E},\tau}$ searches candidate marker poses associated with reference states whose progress lies in $[\widehat{v}_k-\Delta_u,\widehat{v}_k+\Delta_u]$. A candidate physical marker pose is feasible only if it lies on an installable surface, is within the detector range $[r_{\min},r_{\max}]$ from at least one relevant reference pose, lies inside the camera field of view, satisfies the marker-incidence constraint, is not occluded under the map visibility test, and is not a duplicate or near-duplicate of another selected marker pose in workspace. When multiple feasible poses exist, $M_{\mathcal{E},\tau}$ selects the candidate with the highest local visibility score. The realized progress $\widetilde{v}_k$ is the progress value of the nominal reference state at which the selected marker obtains its highest local visibility score. If no feasible candidate exists for any requested progress value, the layout is rejected.

For each task interval $[a_m,b_m]$, define the hard interval count
\begin{equation}
C_m(\widetilde{\mathbf{v}})=\sum_{k=1}^{K}\mathds{1}\!\left[\widetilde{v}_k\in[a_m,b_m]\right].
\end{equation}
The final validation accepts a generated layout only if
\begin{equation}
C_m(\widetilde{\mathbf{v}})\geq n_m\quad\forall m,\qquad |\widetilde{v}_i-\widetilde{v}_j|\geq\delta_{\mathrm{check}}\quad\forall i<j,
\end{equation}
and the physical layout returned by $M_{\mathcal{E},\tau}$ satisfies the installability, visibility, occlusion, range, orientation, and workspace-duplicate constraints described above. We set $\delta_{\mathrm{check}}=\delta_{\min}$ unless otherwise stated.

Before sampling, we reject interval specifications that are trivially infeasible in progress space, such as requiring more separated fiducials than the budget can provide in pairwise disjoint intervals. This check does not prove physical feasibility because physical feasibility also depends on the map, candidate installable surfaces, visibility, and occlusion constraints. If no generated candidate passes the final feasibility check after $R$ samples, we continue sampling up to $R_{\max}$. If no candidate passes after $R_{\max}$ samples, we return the physically feasible candidate with the smallest hard validation violation
\begin{equation}
H(\widetilde{\mathbf{v}})=\sum_{m=1}^{M}\left[n_m-C_m(\widetilde{\mathbf{v}})\right]_++\sum_{i<j}\left[\delta_{\mathrm{check}}-|\widetilde{v}_i-\widetilde{v}_j|\right]_+.
\end{equation}
If no physically feasible candidate exists, the layout is marked infeasible and the fallback baseline specified in the experimental protocol is used.

By construction, any retained candidate satisfies the hard interval-coverage and progress-space separation constraints, and satisfies physical feasibility with respect to the installability and visibility model implemented by $M_{\mathcal{E},\tau}$.

\subsection{Repeated-Sampling Acceptance Probability}

Let $A_r$ be the event that the $r$-th generated candidate passes the final feasibility check. If candidates are generated independently from independent initial Gaussian noise samples and $\mathbb{P}(A_r)=p_{\mathrm{acc}}$ for all $r=1,\ldots,R$, then the probability that at least one of $R$ generated candidates is feasible is
\begin{equation}
\mathbb{P}\!\left(\bigcup_{r=1}^{R}A_r\right)=1-\mathbb{P}\!\left(\bigcap_{r=1}^{R}A_r^c\right)=1-\prod_{r=1}^{R}\mathbb{P}(A_r^c)=1-(1-p_{\mathrm{acc}})^R.
\end{equation}
If candidates have different acceptance probabilities $p_r$ but remain independent, the expression becomes
\begin{equation}
1-\prod_{r=1}^{R}(1-p_r).
\end{equation}
If candidates are generated adaptively, or if rejection of one candidate changes the generation distribution for later candidates, the independence expression need not hold exactly.

\subsection{Implementation Recipes}
\label{Implementation Recipes}

\subsubsection{Dataset Generation}

The training dataset is generated using a simulation-based procedure that varies the environment, disturbance conditions, and fiducial landmark configurations. We first generate a set of mapped environments by sampling obstacle layouts, including the number, type, and placement of obstacles. For each environment, start and goal locations are selected, and a collision-free reference path is planned using the A* algorithm. The planner searches over a discretized representation of the environment to find a shortest feasible path from start to goal while avoiding obstacles. In total, the dataset contains 20 unique maps and planned trajectories.

For each map, we generate 40 different wind fields. Each wind field corresponds to a different disturbance region relative to the path. For every unique map--wind pair, we then generate 50 different fiducial landmark configurations. The landmark configurations are generated randomly using a heuristic constraint-based sampling procedure. Each configuration consists of a fixed number of fiducial tags placed along the planned path.

The heuristic constraints are designed to produce landmark placements that are useful for navigation rather than arbitrary samples along the path. First, a minimum pairwise distance is enforced between landmarks to prevent tags from clustering too closely together. Second, coverage constraints encourage landmarks to be placed in task-relevant regions, including the disturbance region and a subsequent recovery interval, defined as a short segment after the disturbance region where additional localization can help the robot return to the planned path. Additional landmarks are encouraged near obstacle-related regions, such as before obstacles and around obstacle-induced turns, since these portions of the path often correspond to sharper maneuvers and higher tracking uncertainty. We also prevent landmarks from being placed immediately at the start of the trajectory, since the robot begins with a known initial condition and does not benefit significantly from an observation at the starting location.

Each landmark configuration is evaluated using the FalconGym simulator. For every map--wind--landmark configuration, we run 5 stochastic Monte Carlo trials, where each trial uses a different realization of the process noise. During each rollout, we record the environment, the true robot trajectory, whether the robot reaches the goal, and the completion time. The statistics saved for each landmark configuration include the success rate over the 5 trials and the mean completion time.

If the robot fails to reach the goal in a trial, the trial is assigned the maximum allowed completion time. This penalty ensures that failed or poor landmark configurations are represented as low-quality examples during training. The resulting dataset therefore associates each landmark configuration with both its geometric placement and its empirical navigation performance under stochastic disturbances.

\subsubsection{Training Details}
\label{app:training_details}

\begin{figure}[h!]
    \centering
    \includegraphics[width=0.95\linewidth]{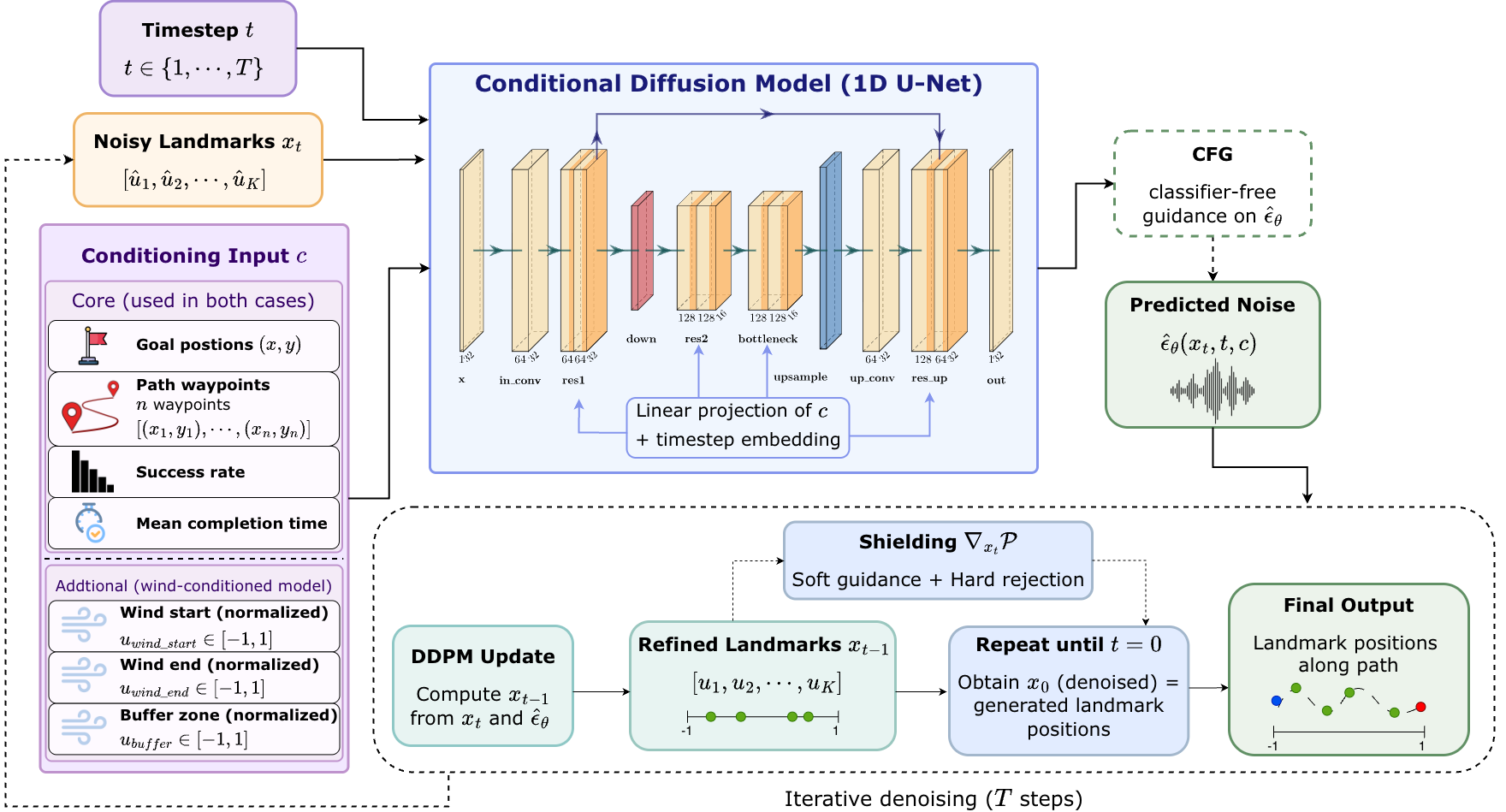}
    \caption{Architecture of the conditional 1D diffusion model used by SCoDA. The denoiser operates on trajectory-progress fiducial layouts and is conditioned on the task context through a learned context projection added to the timestep embedding.}
    \label{fig:scoda_unet_architecture}
\end{figure}

This section gives implementation details for training the conditional diffusion model used in \tbp{SCoDA}. Each training example consists of a candidate fiducial layout, its task context, and rollout statistics obtained by executing the corresponding reference trajectory in simulation. As in Sec.~\ref{subsec:conditional_diffusion}, the diffusion model operates in trajectory-progress space rather than directly in workspace coordinates. Figure ~\ref{fig:scoda_unet_architecture} summarizes the denoising architecture.

\paragraph{Fiducial layout representation.}
For each candidate layout, the $K$ fiducial locations are projected onto the reference trajectory $\omega$ and represented by progress variables $u_i\in[0,1]$, where $u_i=0$ corresponds to the start of the trajectory and $u_i=1$ corresponds to the terminal region. Before diffusion training, these values are linearly scaled to the diffusion range
\begin{equation}
    u^{\mathrm{diff}}_i = 2u_i - 1 \in [-1,1].
\end{equation}
The resulting vector is used as the clean diffusion sample $\mathbf{x}_0$. Fiducial entries are sorted by trajectory progress before training. When padded fiducial slots are used, we store a binary mask and apply the diffusion loss only to valid entries.

\paragraph{Context encoding.}
The conditional input follows the notation in Sec.~\ref{subsec:conditional_diffusion}. The full context is
\begin{equation}
    \mathbf{c}
    =
    [g_x,g_y,\zeta(\omega),\mathbf{d},\mathbf{y}],
\end{equation}

We train two conditioning variants to isolate the effect of disturbance awareness. The base model omits the disturbance interval and uses
\begin{equation}
    \mathbf{c}_{\mathrm{base}}
    =
    [g_x,g_y,\zeta(\omega),\mathbf{y}],
\end{equation}
while the disturbance-conditioned model uses
\begin{equation}
    \mathbf{c}_{\mathrm{dist}}
    =
    [g_x,g_y,\zeta(\omega),\mathbf{d},\mathbf{y}].
\end{equation}
Here,
\begin{equation}
    \mathbf{d}=[u_{d,s},u_{d,e},u_{r,e}],
\end{equation}
where $u_{d,s}$ and $u_{d,e}$ denote the start and end of the disturbance interval in trajectory-progress coordinates, and $u_{r,e}$ denotes the end of the subsequent recovery interval. These interval values are scaled to $[-1,1]$ before being concatenated with the other context features. The base model therefore learns from trajectory geometry and rollout outcomes alone, while the disturbance-conditioned model is given explicit information about where along the trajectory the disturbance occurs and where recovery may be useful.

\paragraph{Timestep and context conditioning.}
For each diffusion timestep $t$, the model uses a learned timestep embedding of dimension 128 followed by a two-layer MLP. The context vector $\mathbf{c}$ is projected with a linear layer into the same 128-dimensional embedding space. The timestep and context embeddings are added to form a combined embedding, which is injected into every residual block through a learned linear projection and broadcast addition over the sequence dimension. This makes the denoiser both time-conditioned and task-conditioned at all stages. We use this additive conditioning pathway rather than cross-attention or FiLM-style scale-shift modulation.
\begin{figure}
    \centering
    \includegraphics[width=0.75\linewidth]{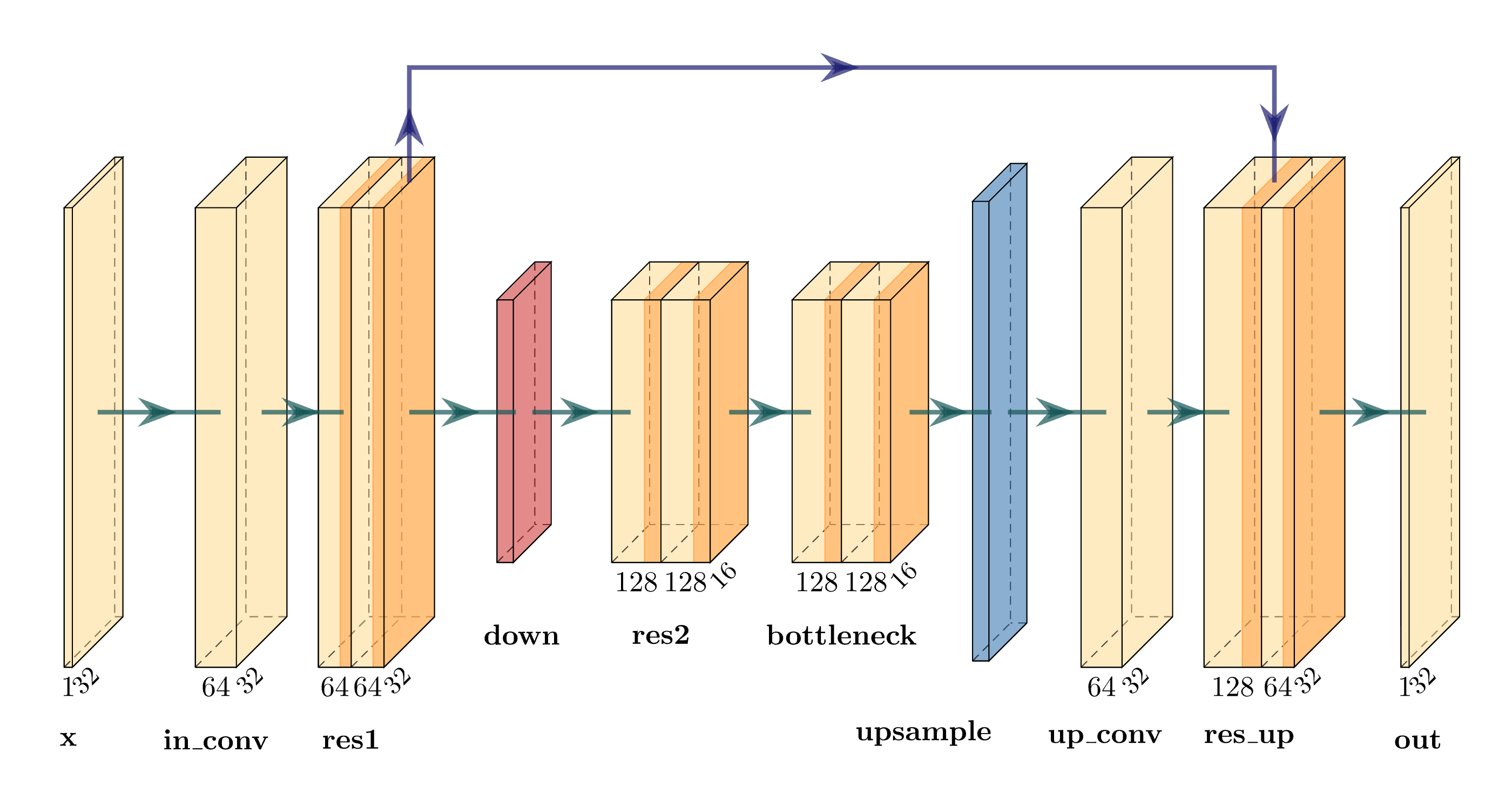}
    \caption{Architecture of the compact conditional 1D U-Net denoiser.}
    \label{fig:1d_unet}
\end{figure}
\paragraph{1D U-Net denoiser.} The architecture of our conditional 1D U-Net denoiser is shown in Figure~\ref{fig:1d_unet}. The denoising network is a compact 1D U-Net that operates on a sequence of fiducial progress values. The input has shape $B\times 1\times T$, where $T$ is the maximum number of fiducial slots. In our experiments, $T=K_{\max}$ and is typically small. The network first applies a 1D convolution from 1 channel to 64 channels with kernel size 3 and padding 1. This is followed by a residual block at the original sequence length. The encoder then downsamples using a stride-2 1D convolution from 64 to 128 channels with kernel size 4 and padding 1, reducing the sequence length from $T$ to $T/2$. A second residual block operates at 128 channels, followed by a 128-channel bottleneck residual block.

The decoder upsamples the bottleneck features by a factor of two using linear interpolation followed by a 1D convolution from 128 channels to 64 channels. The upsampled features are concatenated with the corresponding encoder features through a U-Net skip connection, producing a 128-channel tensor at the original sequence length. A decoder residual block maps this tensor back to 64 channels. Finally, a 1D convolution with kernel size 3 and padding 1 maps the features to a single output channel, producing a noise prediction with the same shape as the input diffusion sample. The final layer uses a $\tanh$ nonlinearity in our implementation to keep the network output numerically bounded.

\paragraph{Residual block.}
Each residual block consists of two 1D convolutional layers with SiLU activations. The combined timestep-context embedding is projected to the block's channel dimension and added after the first convolution, with the embedding broadcast over the sequence dimension. A residual connection adds the block input to the output; when the input and output channel dimensions differ, a $1\times1$ convolution is used on the skip path to match dimensions. This design lets each block adapt its denoising computation to the diffusion timestep and task context while preserving local structure over the ordered fiducial slots.

\paragraph{Diffusion objective.}
Training follows the noise-prediction DDPM objective in Sec.~\ref{subsec:conditional_diffusion}. At each iteration, a diffusion timestep $t$ is sampled uniformly, Gaussian noise $\epsilon$ is added to the clean fiducial vector $\mathbf{x}_0$, producing the noisy sample $\mathbf{x}_t$, and the denoising network is trained to predict the injected noise. For masked layouts, we minimize the masked noise-prediction loss
\begin{equation}
    \mathcal{L}_{\mathrm{train}}
    =
    \frac{
    \sum \left((\epsilon_\theta(\mathbf{x}_t,t,\tilde{\mathbf{c}})-\epsilon)^2 \odot m\right)
    }{
    \sum m + \varepsilon
    },
\end{equation}
where $\epsilon_\theta$ is the predicted noise, $\epsilon$ is the sampled Gaussian noise, $m$ is the binary validity mask over fiducial slots, and $\varepsilon$ is a small constant used for numerical stability.

\paragraph{Classifier-free conditioning dropout.}
To enable classifier-free guidance at generation time, we use conditioning dropout during training. With probability $p_{\mathrm{uncond}}=0.1$, the context vector is replaced by the null context $\tilde{\mathbf{c}}=\mathbf{0}$. Otherwise, the original context is used, $\tilde{\mathbf{c}}=\mathbf{c}$. This allows the same denoising network to learn both conditional and unconditional noise predictions, which are later combined during sampling.

\subsubsection{Shielding Details}
\label{app:shield_details}

We implement shielding at inference time using a two-stage strategy: (1) differentiable soft guidance during denoising, and (2) hard feasibility checks after sampling. The purpose of the shield is to bias the generated fiducial layout toward task-relevant trajectory regions while avoiding redundant placements under the finite marker budget.

During reverse diffusion, the current diffusion state $\mathbf{x}t$ is mapped to trajectory-progress coordinates $\mathbf{v}_t(\mathbf{x}_t)\in[0,1]^K$. Let
\begin{equation}
    \mathcal{I}(\mathbf{c})=\left\{[a_m,b_m]\right\}_{m=1}^{M}
\end{equation}
denote the task-relevant trajectory intervals specified by the context $\mathbf{c}$. In our implementation, these intervals include the disturbance interval $[u_{d,s},u_{d,e}]$, the subsequent recovery interval $[u_{d,e},u_{r,e}]$, and a terminal-approach interval near the goal. The recovery interval corresponds to the short segment after the disturbance region where an additional pose correction may help the robot return to the reference trajectory.

For an interval $[a,b]$, we use the same smooth membership score as in Sec.\ref{subsec:conditional_diffusion},
\begin{equation}
    \psi(v;a,b)
    =
    \sigma(\kappa(v-a))\sigma(\kappa(b-v)),
\end{equation}
which is close to one when $v$ lies inside the interval and close to zero outside it. The shielding objective combines interval coverage and separation penalties as shown in Eq. \eqref{eq:scoda_shield}.

Classifier-free guidance is applied first to form $\hat{\epsilon}_{\mathrm{cfg}}$. The task-gradient is then injected into the predicted noise before the DDPM reverse update as shown in Eq.~\eqref{eq:scoda_shield_update}. As in Sec.\ref{subsec:conditional_diffusion}, shielding is applied only over intermediate reverse steps, after the early noise-dominated states and before the late states become nearly fixed. In our implementation, the guidance window is the middle portion of the reverse process.

After one full denoising pass, the generated vector is clipped to $[-1,1]^K$ and interpreted as the augmentation vector $\mathbf{u}$. For feasibility checking, $\mathbf{u}$ is rescaled to trajectory-progress coordinates in $[0,1]^K$. A sample is accepted only if it satisfies the same interval-coverage and minimum-separation requirements encoded in $J_{\mathrm{task}}$. In particular, each required interval in $\mathcal{I}(\mathbf{c})$ must contain at least one fiducial, and all fiducial pairs must satisfy
\begin{equation}
    |u_i-u_j|
    \geq
    \frac{r_{\mathrm{shield}}}{L_{\mathrm{path}}},
    \quad
    \forall i\neq j,
\end{equation}
where $r_{\mathrm{shield}}$ is the desired physical spacing between fiducials and $L_{\mathrm{path}}$ is the length of the reference trajectory. If any hard check fails, the sample is discarded and a new denoising run is performed, up to a fixed maximum number of attempts.

Accepted samples are sorted by trajectory progress and passed to the deployment map $M_{E,\omega}$, which maps the progress-space augmentation $\mathbf{u}$ to physical fiducial poses in the environment. The soft guidance shapes the reverse diffusion trajectory toward task-relevant regions, while the final hard feasibility check ensures that the returned layout satisfies the specified interval-coverage and separation requirements.
\subsubsection{Hyperparameters}
\label{Hyperparameters}

Table~\ref{tab:scoda_unet_arch} lists each stage of the compact 1D U-Net denoiser. The model operates on a one-dimensional sequence of fiducial-progress values with input shape $B\times 1\times T$, where $T=K_{\max}$. The architecture contains one downsampling stage, one bottleneck residual block, and one upsampling stage with a skip connection from the encoder to the decoder.

\begin{table}[h!]
\centering
\caption{Layer specification of the compact conditional 1D U-Net denoiser. ResBlock1D denotes a residual 1D convolutional block with SiLU activations and additive timestep-context conditioning. The final output has the same sequence length as the input and predicts the diffusion noise for each fiducial slot.}
\small
\setlength{\tabcolsep}{4pt}
\renewcommand{\arraystretch}{1.12}

\begin{tabular}{@{}l!{\vrule width 0.1pt}cccc@{}}
\toprule
{\thcv{Stage}} & {\thcv{In C}} & {\thcv{Out C}} & {\thcv{Operation}} & {\thcv{Output size}} \\
\midrule
\cellcolor{scodaVioletTint}\emph{Input} & & & & \\
Input $\mathbf{x}_t$ & 1 & 1 & -- & $T$ \\
conv & 1 & 64 & Conv1D, $k=3$, pad 1 & $T$ \\
\midrule
\cellcolor{scodaVioletTint}\emph{Encoder} & & & & \\
enc-1 & 64 & 64 & ResBlock1D & $T$ \\
down & 64 & 128 & Conv1D, $k=4$, stride 2, pad 1 & $T/2$ \\
enc-2 & 128 & 128 & ResBlock1D & $T/2$ \\
\midrule
\cellcolor{scodaVioletTint}\emph{Bottleneck} & & & & \\
bottleneck & 128 & 128 & ResBlock1D & $T/2$ \\
\midrule
\cellcolor{scodaVioletTint}\emph{Decoder} & & & & \\
up & 128 & 64 & Interp. $\times 2$ + Conv1D, $k=3$, pad 1 & $T$ \\
skip concat & 64+64 & 128 & Concatenate encoder feature & $T$ \\
dec-1 & 128 & 64 & ResBlock1D & $T$ \\
\midrule
\cellcolor{scodaVioletTint}\emph{Output} & & & & \\
noise head & 64 & 1 & Conv1D, $k=3$, pad 1 + $\tanh$ & $T$ \\
\bottomrule
\end{tabular}
\label{tab:scoda_unet_arch}
\end{table}


\newpage
\section{Additional Qualitative Results}
\label{proofs}

We evaluate \tbp{SCoDA} on unseen test environments that are not included in the training dataset. The test set is constructed by varying the start and goal locations to generate five different maps and planned paths. For each map, we evaluate three wind-region placements along the path: one near the beginning of the trajectory, one near the middle, and one near the goal. This results in fifteen map–wind test scenarios.

\begin{figure}[htbp]
    \centering

    \begin{subfigure}{0.9\textwidth}
        \centering
        \includegraphics[width=\textwidth]{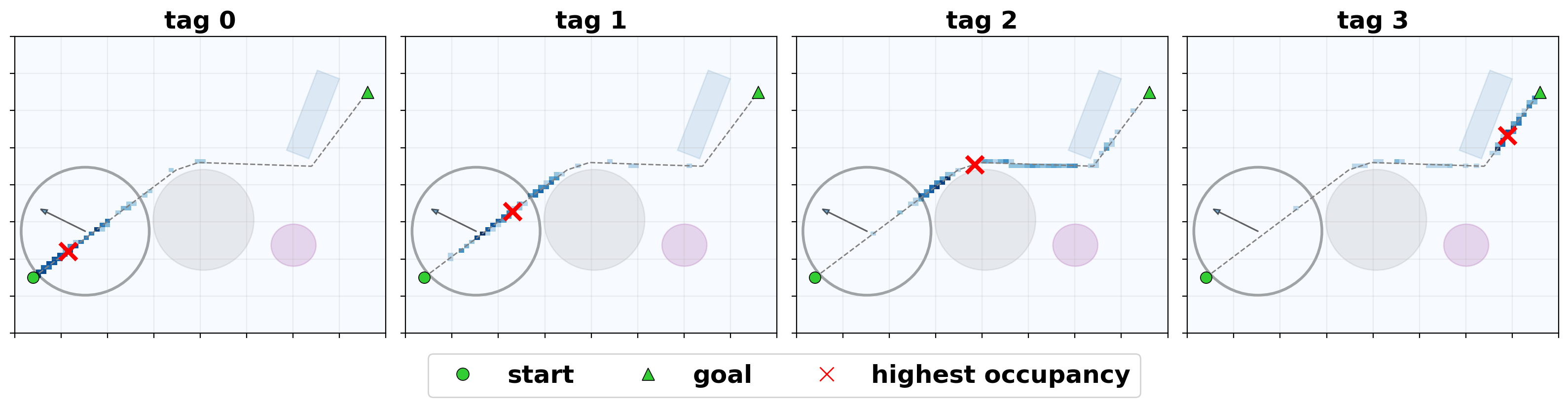}
        \caption{Map 1}
        \label{fig:subplot1}
    \end{subfigure}

    \vspace{0.1cm}

    \begin{subfigure}{0.9\textwidth}
        \centering
        \includegraphics[width=\textwidth]{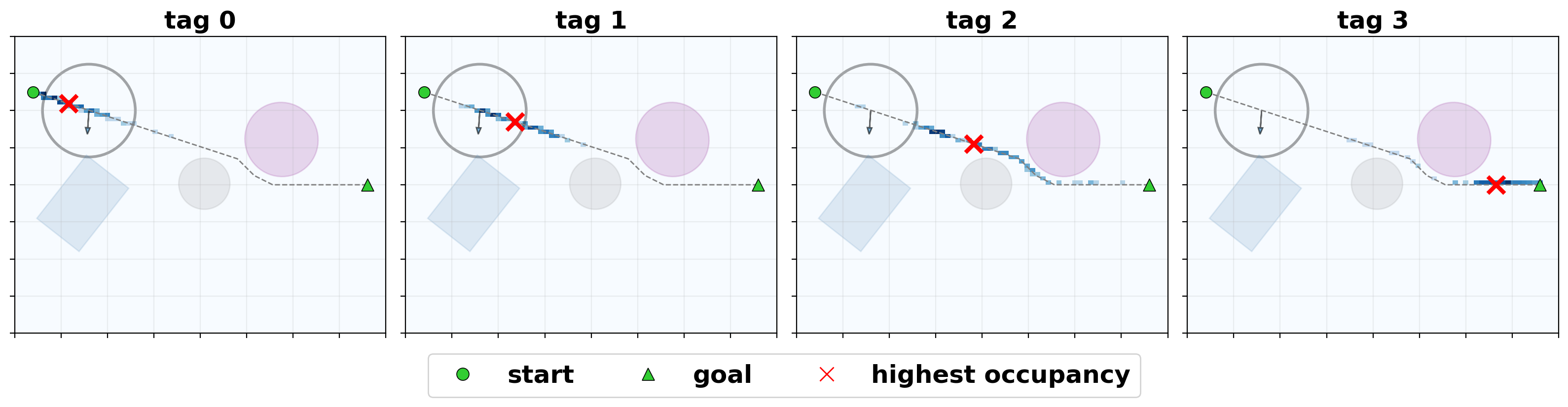}
        \caption{Map 2}
        \label{fig:subplot2}
    \end{subfigure}

    \vspace{0.1cm}

    \begin{subfigure}{0.9\textwidth}
        \centering
        \includegraphics[width=\textwidth]{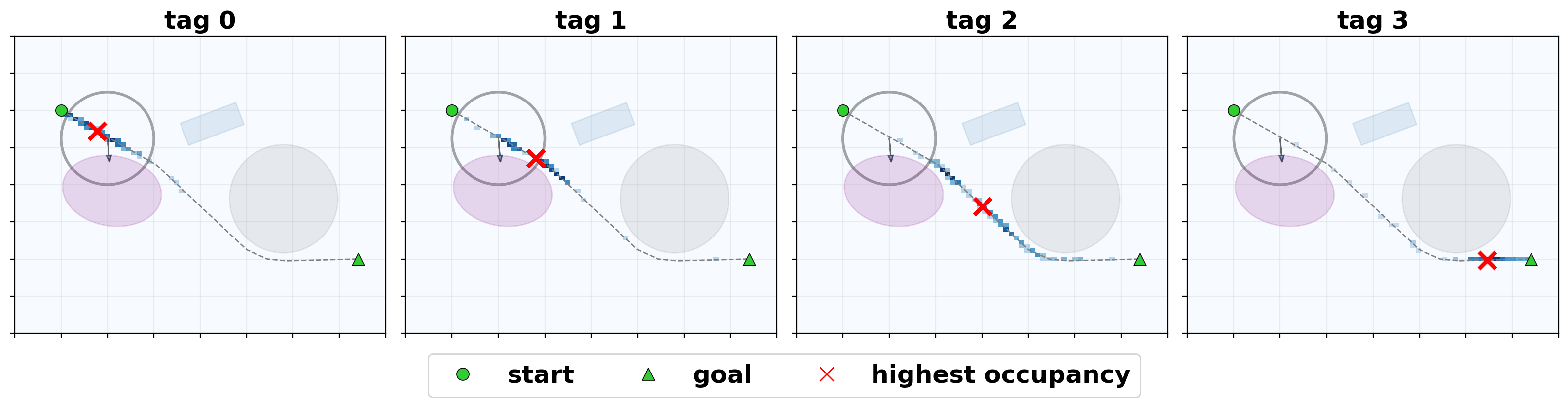}
        \caption{Map 3}
        \label{fig:subplot3}
    \end{subfigure}

    \vspace{0.1cm}

    \begin{subfigure}{0.9\textwidth}
        \centering
        \includegraphics[width=\textwidth]{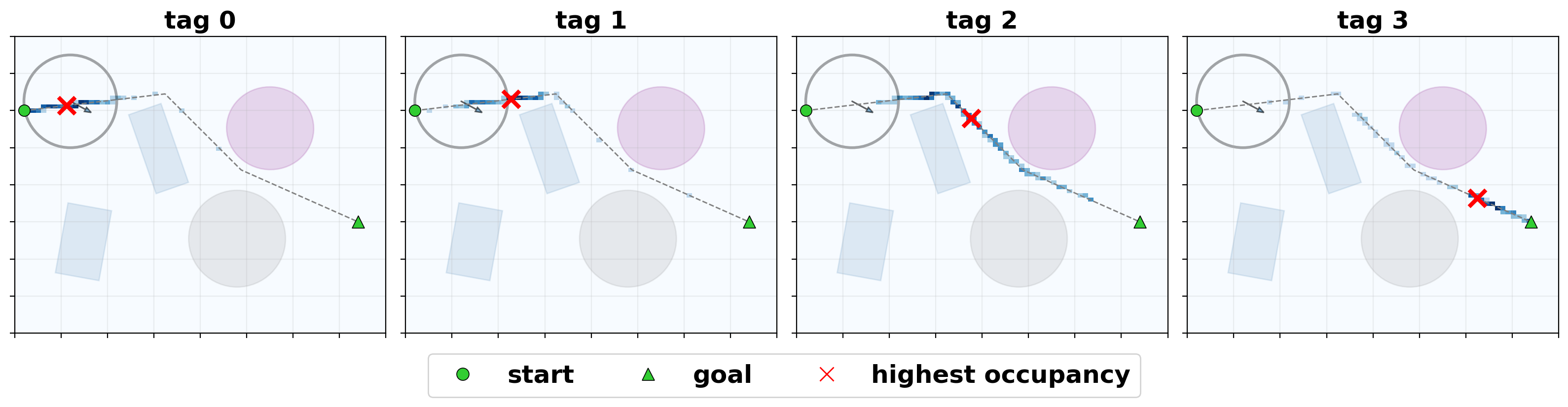}
        \caption{Map 4}
        \label{fig:subplot4}
    \end{subfigure}

    \vspace{0.1cm}

    \begin{subfigure}{0.9\textwidth}
        \centering
        \includegraphics[width=\textwidth]{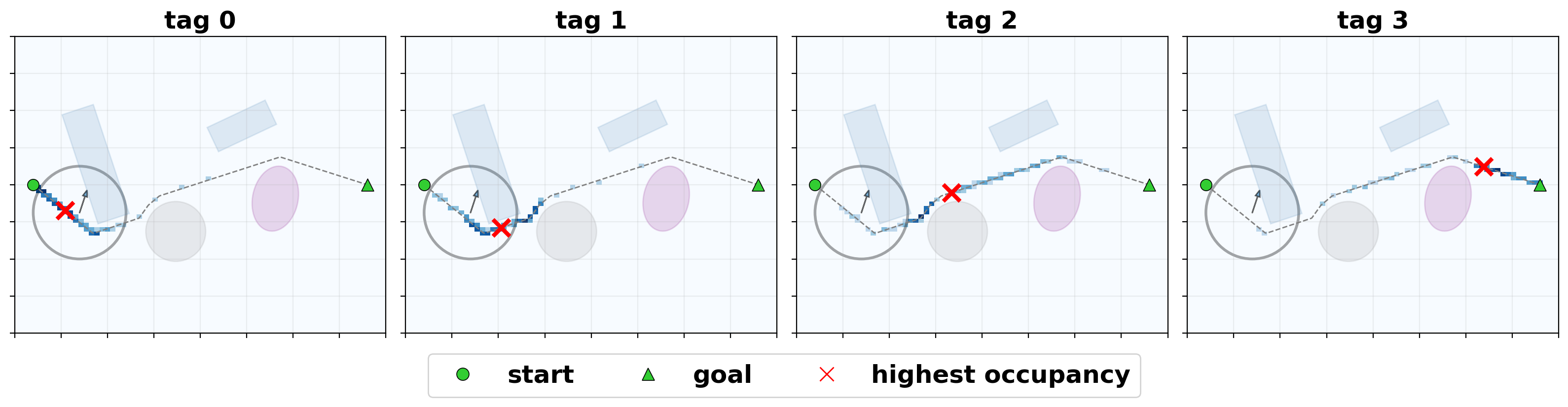}
        \caption{Map 5}
        \label{fig:subplot5}
    \end{subfigure}

    \caption{Top-ranked fiducial layouts across different maps affected by early-path disturbance regions.}
    \label{fig:early}
\end{figure}

\begin{figure}[htbp]
    \centering

    \begin{subfigure}{0.9\textwidth}
        \centering
        \includegraphics[width=\textwidth]{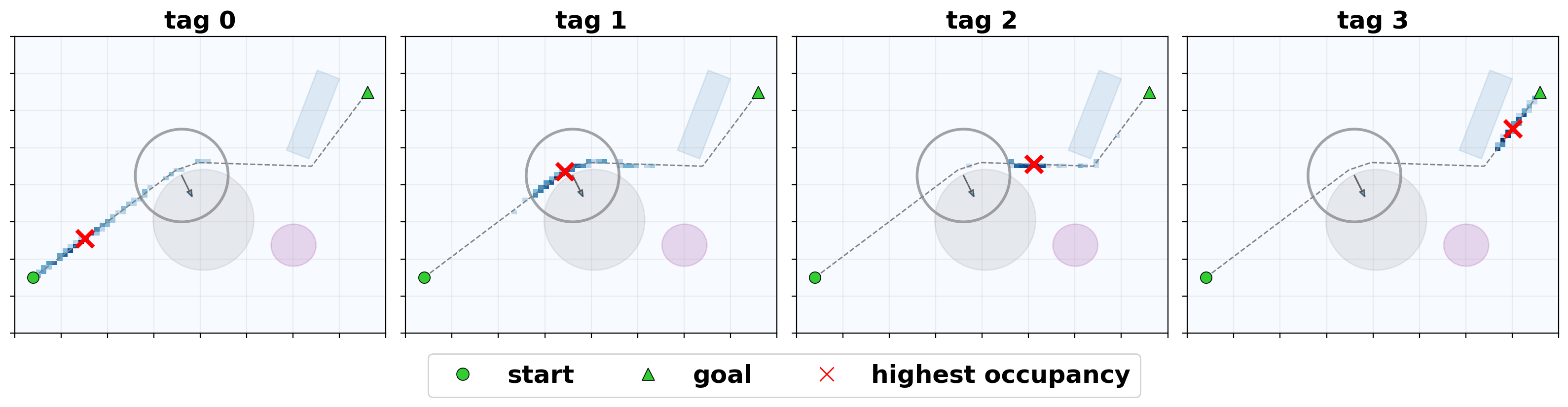}
        \caption{Map 1}
        \label{fig:subplot1}
    \end{subfigure}

    \vspace{0.1cm}

    \begin{subfigure}{0.9\textwidth}
        \centering
        \includegraphics[width=\textwidth]{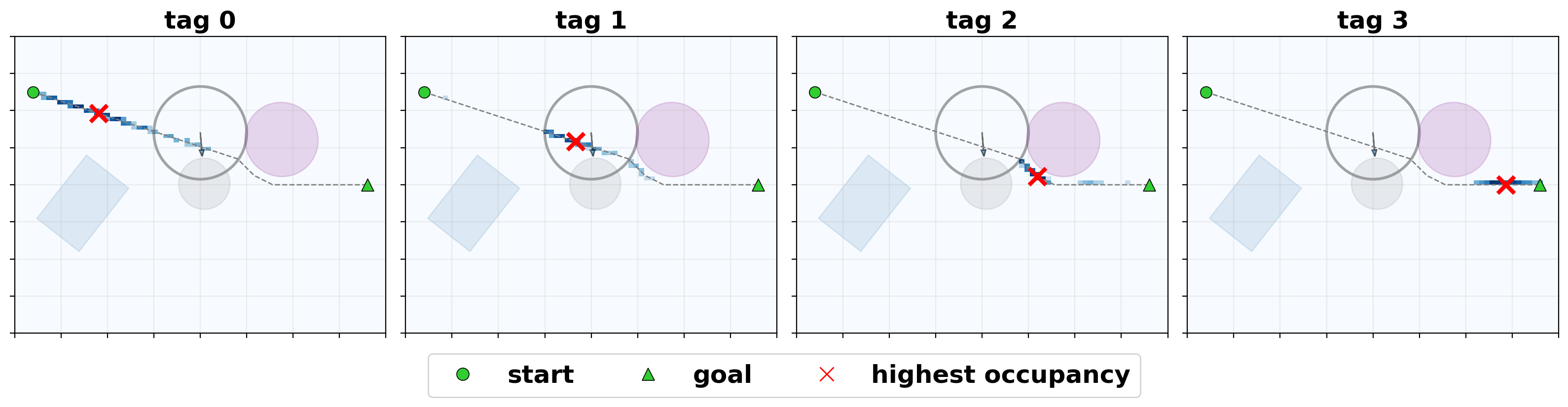}
        \caption{Map 2}
        \label{fig:subplot2}
    \end{subfigure}

    \vspace{0.1cm}

    \begin{subfigure}{0.9\textwidth}
        \centering
        \includegraphics[width=\textwidth]{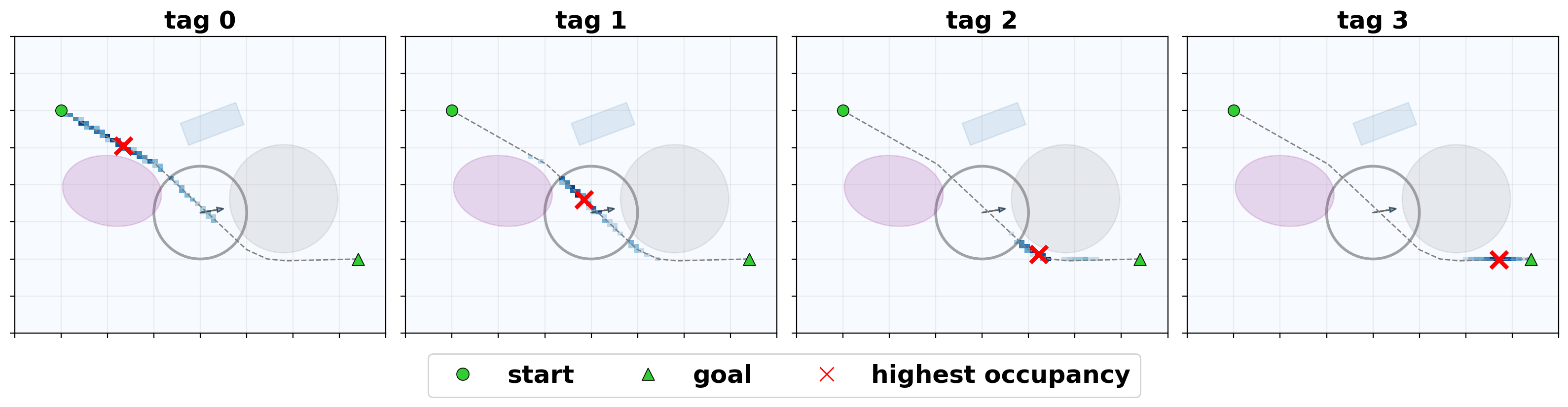}
        \caption{Map 3}
        \label{fig:subplot3}
    \end{subfigure}

    \vspace{0.1cm}

    \begin{subfigure}{0.9\textwidth}
        \centering
        \includegraphics[width=\textwidth]{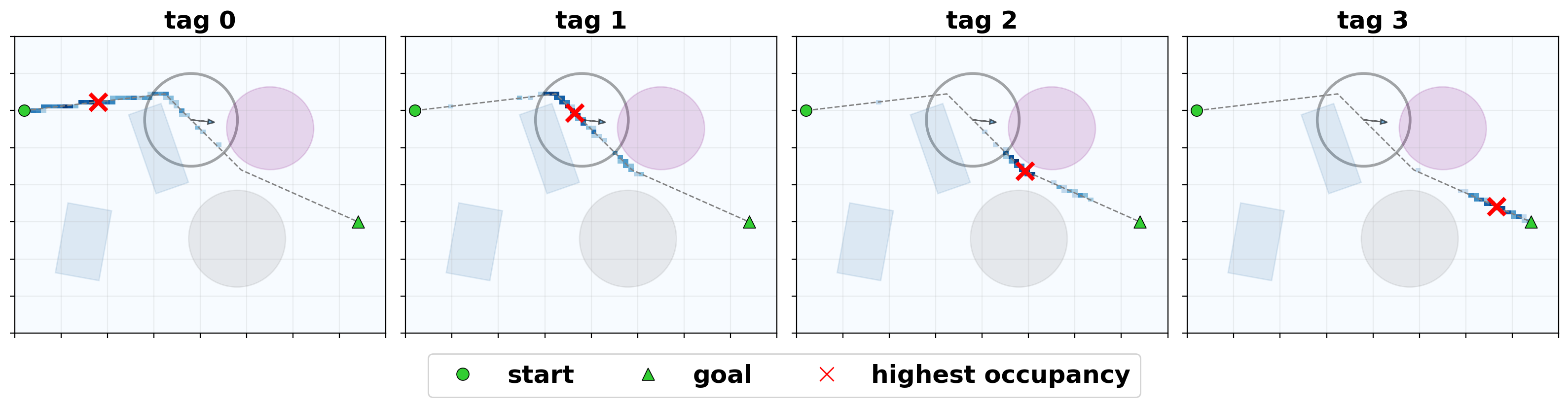}
        \caption{Map 4}
        \label{fig:subplot4}
    \end{subfigure}

    \vspace{0.1cm}

    \begin{subfigure}{0.9\textwidth}
        \centering
        \includegraphics[width=\textwidth]{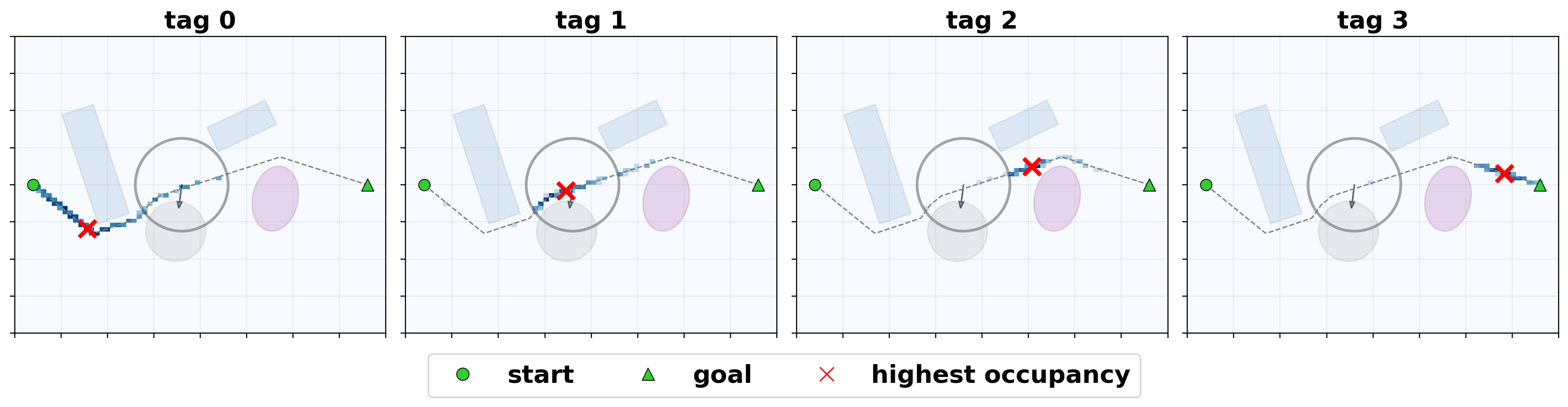}
        \caption{Map 5}
        \label{fig:subplot5}
    \end{subfigure}

    \caption{Top-ranked fiducial layouts across different maps affected by mid-path disturbance regions.}
    \label{fig:five_vertical_subplots}
\end{figure}

\begin{figure}[htbp]
    \centering

    \begin{subfigure}{0.9\textwidth}
        \centering
        \includegraphics[width=\textwidth]{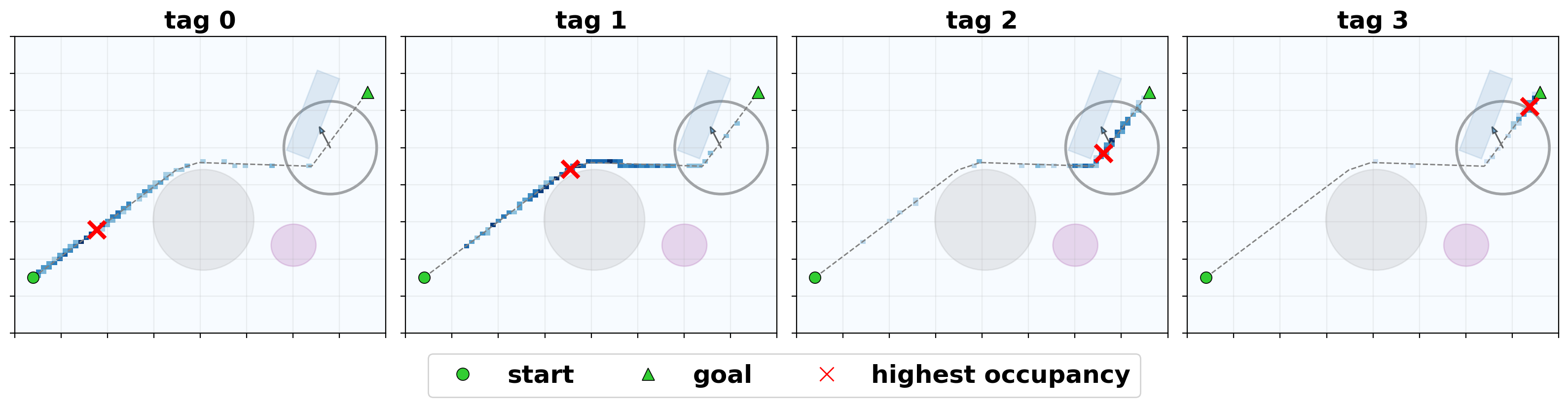}
        \caption{Map 1}
        \label{fig:subplot1}
    \end{subfigure}

    \vspace{0.1cm}

    \begin{subfigure}{0.9\textwidth}
        \centering
        \includegraphics[width=\textwidth]{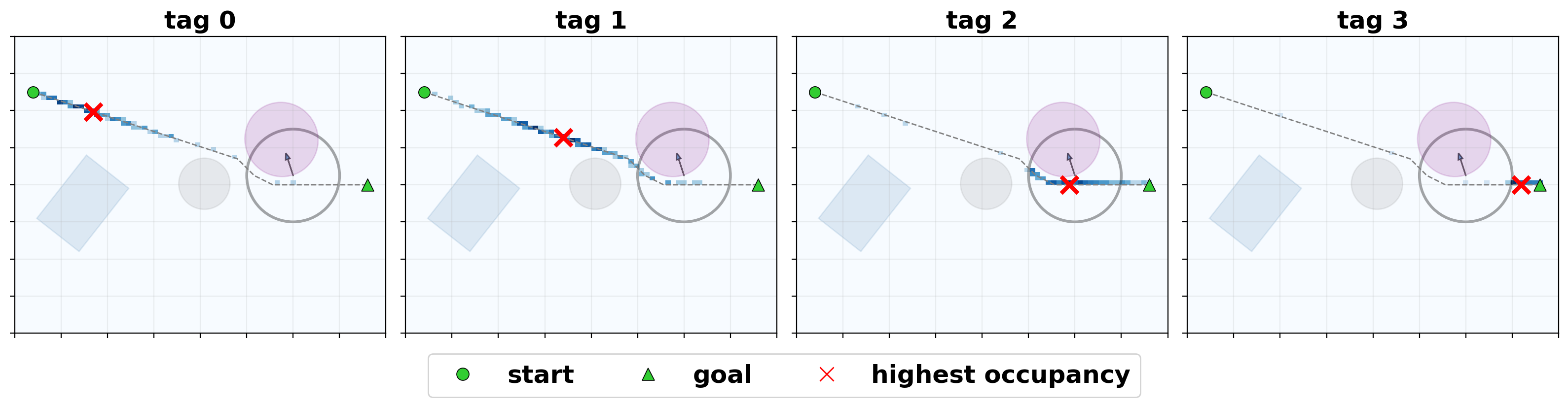}
        \caption{Map 2}
        \label{fig:subplot2}
    \end{subfigure}

    \vspace{0.1cm}

    \begin{subfigure}{0.9\textwidth}
        \centering
        \includegraphics[width=\textwidth]{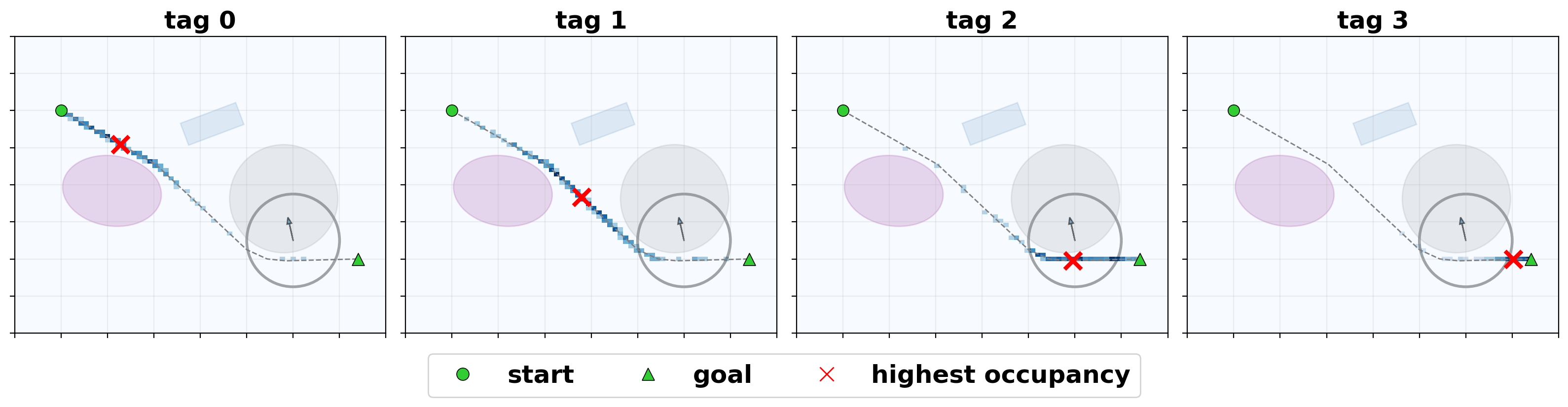}
        \caption{Map 3}
        \label{fig:subplot3}
    \end{subfigure}

    \vspace{0.1cm}

    \begin{subfigure}{0.9\textwidth}
        \centering
        \includegraphics[width=\textwidth]{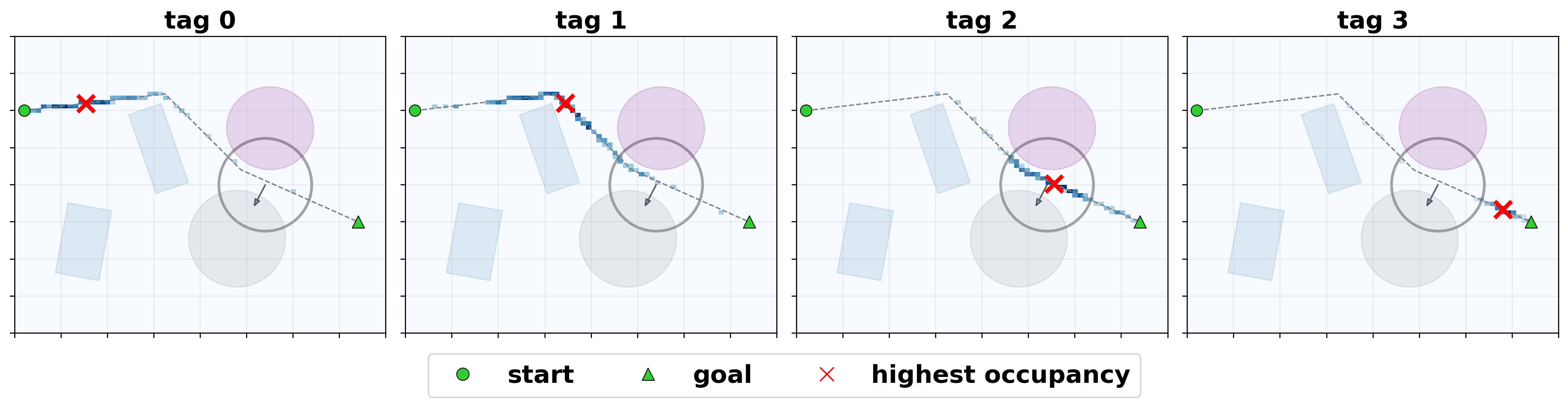}
        \caption{Map 4}
        \label{fig:subplot4}
    \end{subfigure}

    \vspace{0.1cm}

    \begin{subfigure}{0.9\textwidth}
        \centering
        \includegraphics[width=\textwidth]{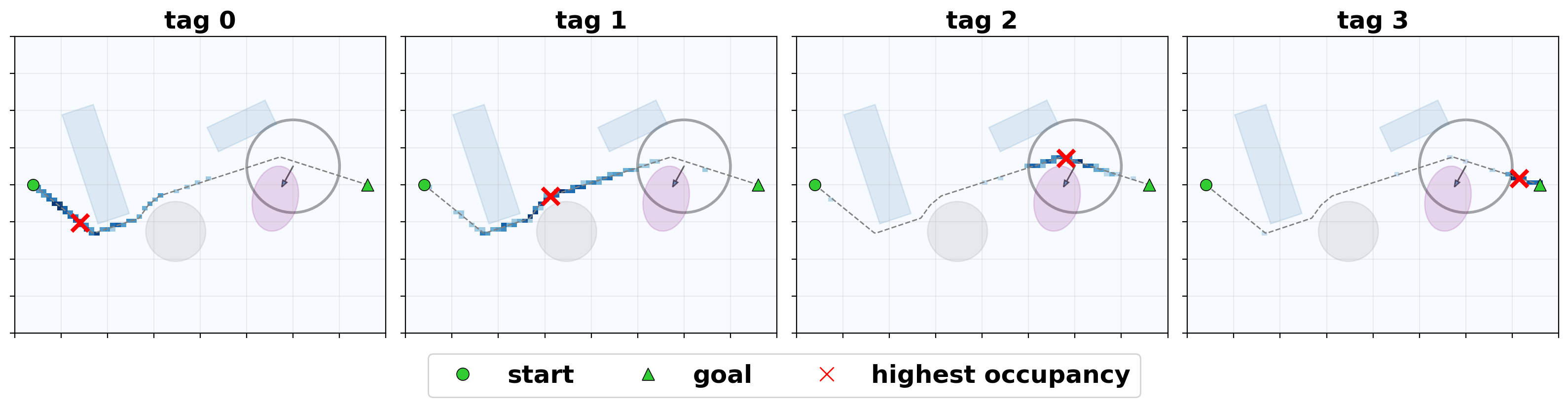}
        \caption{Map 5}
        \label{fig:subplot5}
    \end{subfigure}

    \caption{Top-ranked fiducial layouts across different maps affected by end-of-path disturbance regions.}
    \label{fig:end}
\end{figure}

\paragraph{Qualitative visualization of generated layouts.}
Figures~\ref{fig:early}--\ref{fig:end} show the raw \tbp{SCoDA} outputs for five unseen maps with early-, middle-, and end-of-path disturbance regions. The dashed curve is the planned trajectory, the green markers indicate start and goal, and the shaded shapes denote obstacles and disturbance regions. Each column corresponds to one fiducial slot. For each slot, we draw many generated samples from \tbp{SCoDA} and project the resulting fiducial locations back onto the reference trajectory. The blue heatmap visualizes the empirical occupancy of these samples along the path: darker or denser regions indicate trajectory segments where that fiducial slot is generated more frequently across samples. The red $\times$ marks the highest-occupancy point, corresponding to the mode of the generated distribution for that slot. This visualization therefore shows not only a single selected layout, but also the distributional structure learned by the model. The results illustrate that \tbp{SCoDA} places probability mass near task-relevant regions, such as disturbance intervals and  subsequent recovery segments, obstacle-induced turns, and terminal-approach regions.



\newpage
\subsection{Early-path disturbance regions}

\newpage
\subsection{Mid-path disturbance regions}

\newpage
\subsection{End-of-path disturbance regions}

\end{appendix}

\end{document}